# Best-first Model Merging for Hidden Markov Model Induction

Andreas Stolcke* Stephen M. Omohundro†

**Abstract**

This report describes a new technique for inducing the structure of Hidden Markov Models from data which is based on the general 'model merging' strategy (Omohundro 1992). The process begins with a maximum likelihood HMM that directly encodes the training data. Successively more general models are produced by merging HMM states. A Bayesian posterior probability criterion is used to determine which states to merge and when to stop generalizing. The procedure may be considered a heuristic search for the HMM structure with the highest posterior probability. We discuss a variety of possible priors for HMMs, as well as a number of approximations which improve the computational efficiency of the algorithm.

We studied three applications to evaluate the procedure. The first compares the merging algorithm with the standard Baum-Welch approach in inducing simple finite-state languages from small, positive-only training samples. We found that the merging procedure is more robust and accurate, particularly with a small amount of training data. The second application uses labelled speech data from the TIMIT database to build compact, multiple-pronunciation word models that can be used in speech recognition. Finally, we describe how the algorithm was incorporated in an operational speech understanding system, where it is combined with neural network acoustic likelihood estimators to improve performance over single-pronunciation word models.

*Computer Science Division, University of California at Berkeley, and International Computer Science Institute, 1947 Center Street, Berkeley, CA 94704, e-mail stolcke@icsi.berkeley.edu.

†International Computer Science Institute, 1947 Center Street, Berkeley, CA 94704, e-mail om@icsi.berkeley.edu.

# Contents

# List of Figures

# 1 Introduction and Overview

Hidden Markov Models (HMMs) are a popular method for modeling stochastic sequences with an underlying finite-state structure. Some of their first uses were in the area of crypt-analysis and they are now the model of choice for speech recognition (Rabiner & Juang 1986). Recent applications include part-of-speech tagging (Cutting *et al.* 1992) and protein classification and alignment (Haussler *et al.* 1992; Baldi *et al.* 1993). Because HMMs can be seen as probabilistic generalizations of non-deterministic finite-state automata they are also of interest from the point of view of formal language induction.

For most modeling applications it is not feasible to specify HMMs by hand. Instead, the HMM needs to be at least partly estimated from available sample data. All of the applications mentioned crucially involve *learning*, or adjusting the HMM to such data. Standard HMM estimation techniques assume knowledge of the model size and structure (or topology) and proceed to optimize the continuous model parameters using well-known statistical techniques. Section 2 defines the HMM formalism and gives an overview of these standard estimation methods.

In contrast to traditional HMM estimation based on the Baum-Welch technique (Baum *et al.* 1970), our method uses Bayesian evidence evaluation to adjust the model topology to the data. The approach is based on the idea that models should evolve from simply storing examples to representing more complex and general relationships as increasing amounts of data become available. The transition is gradual and is accomplished by successive merging of submodels. This general approach and its application to HMMs are discussed in Section 3.

As a result of our implementation and applications of the merging algorithm, a number of crucial efficiency improvements, approximations and heuristics have been developed. These are discussed in Section 4.

Our approach is related to ideas that have appeared in the literature, in some cases for considerable time. Section 5 discusses some of these links to related work and compares the various approaches.

The HMM induction-by-merging idea is evaluated experimentally using both artificial and realistic applications in Section 6. We compare the structure-induction capabilities of our method to those of the Baum-Welch method, and find that it produces models that have better generalization and/or are more compact. In particular, applications in the area of phonetic word modeling for speech recognition show that HMM merging can be an effective and efficient tool in practice.

Finally, we draw some conclusions and point to continuations of this work in Section 7.

# 2 Hidden Markov Models

## 2.1 Informal characterization

Hidden Markov Models (HMMs) can be described from at least two perspectives. In one view, they are a stochastic generalization of the nondeterministic finite automata (NFAs) studied in automata theory (*e.g.* Hopcroft & Ullman (1979)). As NFAs, HMMs have a finite number of states, including the initial and final ones. States are connected by transitions and emit output symbols from a finite, discrete alphabet. (HMMs can also be defined to emit

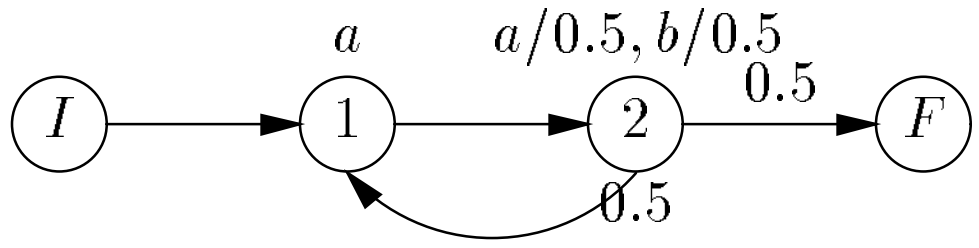

Figure 1: Simple HMM.

State names appear within circles, outputs above their emitting states.

output symbols on the transitions rather than at states, as is usual with NFAs. However, the two variants are equivalent and can be converted into each other.) As NFAs, HMMs generate (or accept) strings over the output alphabet by nondeterministic walks between the initial and final states. In addition, HMMs also assign probabilities to the strings they generate, computed from the probabilities of individual transitions and emissions.

From a probabilistic point-of-view, HMMs characterize a class of stochastic processes that have an underlying Markovian finite-state structure, but one that may only be observed indirectly (hence the "hidden" nature of the model). Specifically, one has a sequence of observable random variables (the emissions) that each depend only on an underlying state variable. The hidden state variables in turn depend only on their immediate predecessors, *i.e.*, they form a Markov chain. We will be discussing only first-order HMMs where state dependencies are on the immediate predecessor only; higher-order HMMs can have states depending on a limited number of predecessors. Higher-order HMMs may be transformed into equivalent first-order ones.

In this kind of setting one sometimes wants to generate unbounded sequences, in which case there are no final states. Instead, under certain reachability conditions, the model settles into a unique stationary distribution of states and outputs (*e.g.*, Cover & Thomas (1991)). Also, there is no reason why the model is limited to discrete outputs. Continuous output distributions are commonly used in modelling speech, for example.

We will primarily adopt the automata-theoretic perspective, describing HMMs as stochastic NFAs. In this context they offer an attractive probabilistic learning framework, in which languages can be learned from representative, random positive samples. Although this paper deals with only discrete-output HMMs, our learning approach can be generalized to other kinds of output distribution, as briefly discussed in Section 7.

### 2.2 An Example

Before introducing formal notation, consider the simple HMM example depicted in Figure 1.

The model generates strings from the regular language $(a(a \cup b))^*$. All transition and emission probabilities are 1 except where labeled otherwise. The exceptions are at state 2, which outputs $a$ or $b$ with probability 0.5 each, and where transitions occur to either state 1 or the final state $F$, again with probability 0.5.

The model not only accepts or rejects strings, it also assign probabilities to strings. The conditional probability $P(x|M)$ of a string $x$ given a model $M$, is the sum of all the joint probabilities of random walks through the model generating $x$.

In the example HMM, the string *abaa* is generated by only a single path through the model, and hence

$$\begin{aligned} P(abaa|M) &= p(q_I \to q_1)p(q_1 \uparrow a)p(q_1 \to q_2)p(q_2 \uparrow b) \\ &\quad p(q_2 \to q_1)p(q_1 \uparrow a)p(q_1 \to q_2)p(q_2 \uparrow a)p(q_2 \to q_F) \\ &= (0.5)^4 = 0.1375. \end{aligned}$$

The conditional probabilities in the product are the individual transition and emission probabilities. The fact that each is conditional only on the current state reflects the Markovian character of the generation process.

## 2.3 Definitions

We now define these concepts more formally.[1] A *(discrete output, first-order) Hidden Markov Model* is specified by a set of *states* $Q$, an *output alphabet* $\Sigma$, an *initial state* $q_I$, a *final state* $q_F$, and a set of probability parameters. *Transition probabilities* $p(q \to q')$ specify the probability that state $q'$ follows $q$, for all $q, q' \in Q$. *Emission (output) probabilities* $p(q \uparrow \sigma)$ specify the probability that symbol $\sigma$ is emitted while in state $q$, for all $q \in Q$ and $\sigma \in \Sigma$.

By the *structure* or *topology* of an HMM we mean its states $Q$, its outputs $\Sigma$, a subset of its transitions $q \to q'$ with $p(q \to q') = 0$ and a subset of its emissions $q \uparrow \sigma$ with $p(q \uparrow \sigma) = 0$. In other words, an HMM topology specifies a subset of the potential transitions and emissions which are guaranteed to have zero probability, and leaves the remaining probabilities unspecified.

We use superscripts on states $q^t$ and emissions $\sigma^t$ to denote discrete time indices in the generation of an output sequence. Therefore,

$$p(q \to q') = p((q')^{t+1}|q^t), \quad t = 0, 1, 2, \ldots$$

and

$$p(q \uparrow \sigma) = p(\sigma^t|q^t), \quad t = 0, 1, 2, \ldots$$

The initial state $q_I$ occurs at the beginning of any state sequence and the final state $q_F$ at the end of any complete state sequence. Neither $q_I$ nor $q_F$ can occur anywhere else, and they do not emit symbols. For convenience we assume $q_I, q_F \notin Q$.

An HMM is said to *generate* a string $x = x_1 x_2 \ldots x_\ell \in \Sigma^*$ if and only if there is a state sequence, or *path*, $q_1 q_2 \ldots q_\ell \in Q^*$ with non-zero probability, such that $q_t$ outputs $x_t$ with non-zero probability, for $t = 1, \ldots, \ell$. The *probability of a path* (relative to $x$) is the product of all transition and emission probabilities along it.

The *conditional probability* $P(x|M)$ of a string $x$ given an HMM $M$ is computed as the sum of the probabilities of all paths that generate $x$:

$$P(x|M) = \sum_{q_1 \ldots q_\ell \in Q^\ell} p(q_I \to q_1)p(q_1 \uparrow x_1)p(q_1 \to q_2) \ldots p(q_\ell \uparrow x_\ell)p(q_\ell \to q_F) \tag{1}$$

[1] Where possible we try to keep the notation consistent with Bourlard & Morgan (1993).

## 2.4 HMM estimation

The Baum-Welch estimation method for HMMs (Baum *et al.* 1970) assumes a certain topology and adjusts the parameters so as to maximize the model likelihood on the given samples. If the structure is only minimally specified (*i.e.*, all probabilities can assume non-zero values) then this method can potentially *find* HMM structures by setting a subset of the parameters to zero (or close enough to zero so that pruning is justified).

The fundamental problem in HMM estimation is that the state variables are not directly observable. If they were, *i.e.*, if we could observe sequences of states $q_1 q_2 \ldots q_\ell$ in addition to the outputs, estimation of the probability parameters would be straightforward. One could collect sufficient statistics,

$$\begin{aligned} c(q \to q') &= \text{number of transitions from state } q \text{ to } q', \text{ for all } q, q' \in \mathcal{Q} \\ c(q \uparrow \sigma) &= \text{number of outputs of } \sigma \text{ from state } q, \text{ for all } q \in \mathcal{Q}, \sigma \in \Sigma, \end{aligned}$$

and set the model parameters to their maximum likelihood values:

$$\hat{p}(q \to q') = \frac{c(q \to q')}{\sum_{s \in \mathcal{Q}} c(q \to s)} \tag{2}$$

$$\hat{p}(q \uparrow \sigma) = \frac{c(q \uparrow \sigma)}{\sum_{\rho \in \Sigma} c(q \uparrow \rho)}. \tag{3}$$

The problem of the missing state observations can be solved by replacing the unknown transition and output frequencies by their expected values given a current model estimate and the sample output sequences. For each sample sequence $x$ we compute the posterior probability $P(q^t|x, M)$ that the path generating $x$ passes through state $q$ at time $t$. This can be done by an efficient $O(\ell|\mathcal{Q}|^2)$ dynamic programming algorithm know as the *forward-backward* algorithm. From $P(q^t|x, M)$ for all $q$ and $t$, it is then straightforward to compute the posterior expectations

$$\begin{aligned} \hat{c}(q \to q') &= E[c(q \to q')|X, M] \\ \hat{c}(q \uparrow \sigma) &= E[c(q \uparrow \sigma)|X, M]. \end{aligned}$$

The model parameters are then maximized with respect to the expectations $\hat{c}$ instead of the unknown values $c$. Re-estimating parameters affects the expectations, so $\hat{c}$ has to be recomputed, parameters estimated again, *etc.*, until a fixed point is reached. This iterative procedure is a special case of the general EM (expectation-maximization) method for estimating distributions with hidden parameters (Dempster *et al.* 1977).

However, the method is not fool-proof: since it uses what amounts to a hill-climbing procedure that is only guaranteed to find a local likelihood maximum, the result of Baum-Welch estimation may turn out to be sub-optimal. In particular, results will depend on the initial values chosen for the model parameters. Several examples of this phenomenon will be seen in Section 6.1.

## 2.5 Viterbi approximation

A frequently used approximation in HMM estimation is to proceed as if each sample comes from only a single path through the model, *i.e.*, all paths except the most likely one are assumed to have zero or negligible probability. The most likely, or *Viterbi path* (after Viterbi (1967)) is the one that maximizes the summand in equation (1):

$$V(x|M) = \operatorname*{argmax}_{q_1 \ldots q_\ell \in \mathcal{Q}^\ell} p(q_I \to q_1) p(q_1 \uparrow x_1) p(q_1 \to q_2) \ldots p(q_\ell \uparrow x_\ell) p(q_\ell \to q_F) \qquad (4)$$

Let $V_i(x|M)$ denote the $i$th state in $V(x|M)$, with $V_0(x|M) = q_I$ and $V_\ell(x|M) = q_F$ for convenience.

By neglecting all paths except $V(x|M)$, the statistics used in re-estimating transitions and emissions become sums of 0's and 1's. The resulting approximated estimates are

$$\begin{aligned} \hat{c}(q \to q') &= \sum_{x \in X} \sum_{i=0}^{\ell} \delta(q, V_i(x|M)) \delta(q', V_{i+1}(x|M)) \\ \hat{c}(q \uparrow \sigma) &= \sum_{x \in X} \sum_{i=1}^{\ell} \delta(q, V_i(x|M)) \delta(\sigma | x_i), \end{aligned}$$

where the Kronecker delta $\delta(x, y)$ is 1 if $x = y$ and 0 otherwise.

We mention the Viterbi approximation here because it also turns out to be very useful in an efficient implementation of the HMM induction algorithm described in later sections.

# 3 HMM Induction by Bayesian Model Merging

## 3.1 Model merging

The approach to HMM induction presented here was motivated by earlier work by one of us (Omohundro 1992) on *model merging* as a fundamental, cognitively plausible induction technique that is applicable to a variety of domains. The basic idea is that a model of a domain is constructed from submodels. When only a small amount of data is available, the submodels consist of the data points themselves with similarity-based generalization. As more data becomes available, submodels from more complex families may be constructed. The elementary induction step is to successively *merge* pairs of simple submodels to form more complex submodels. Because the new submodel is based on the combined data from the merged submodels, it may be reliably chosen from a more complex model space. This approach allows arbitrarily complex models to be constructed without overfitting. The resulting models adapt their representational power to the structure of the training data.

The search for submodels to merge is guided by an attempt to sacrifice as little of the sample likelihood as possible as a result of the merging process. This search can be done very efficiently if (a) a greedy search strategy can be used, and (b) likelihood computations can be done locally for each submodel and don't require global recomputation on each model update.

This idea can be intuitively illustrated in a geometric domain. Consider the problem of modeling a curve in the plane by a combination of straight line segments. The likelihood in this case corresponds to the mean square error from each curve point to the nearest segment point.[2] A merging step in this case consists of replacing two segments by a single segment.

[2]More precisely, the mean squared error is proportional to the log probability of the data under the assumption that each curve generates points normally distributed around them.

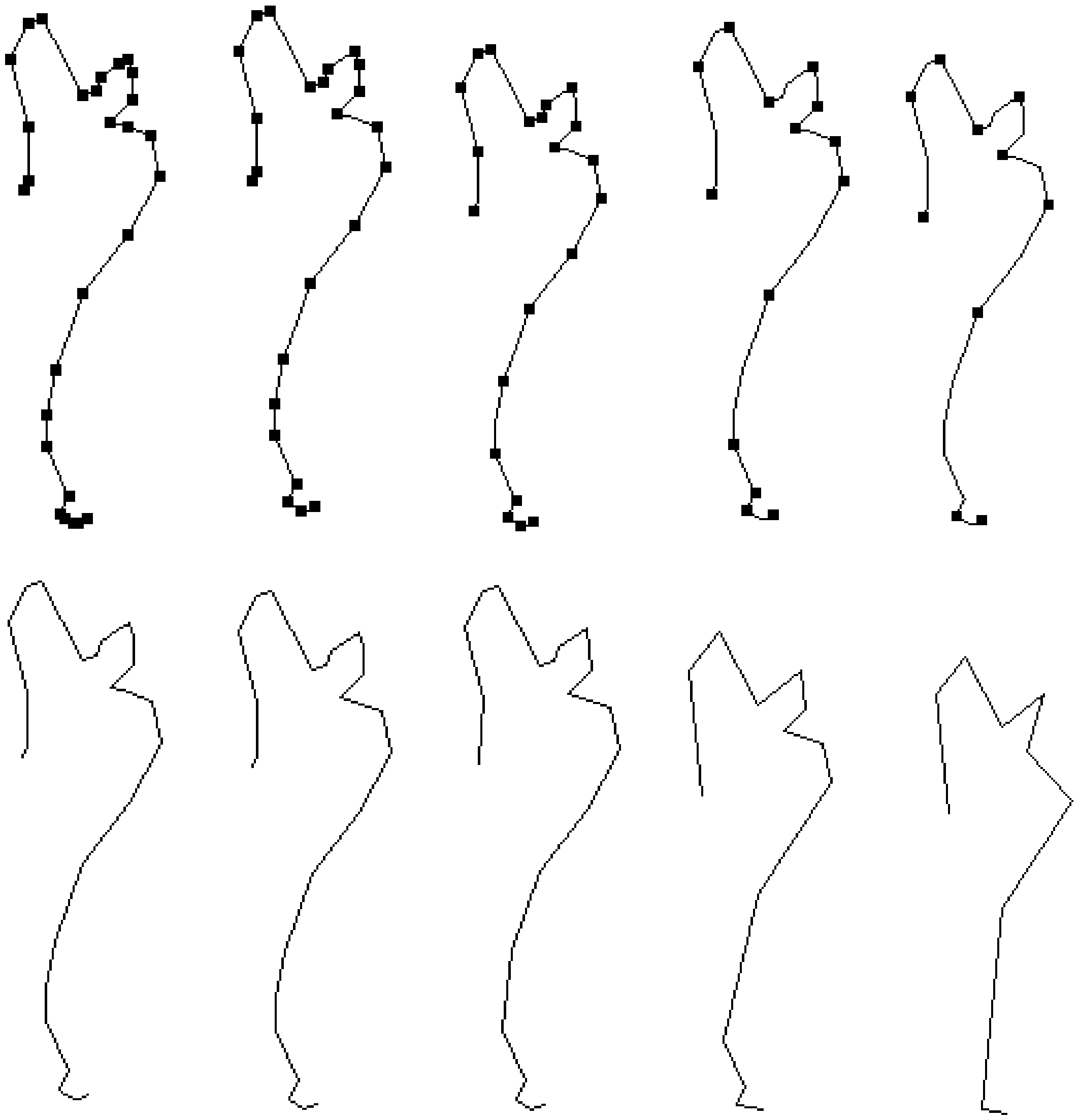

Error = 1 Error = 2 Error = 5 Error = 10 Error = 20

Figure 2: Approximation of a curve by best-first merging of segment models.

The top row shows the endpoints chosen by the algorithm at various levels of allowed error. The bottom row shows the corresponding approximation to the curve.

We always choose that pair such that the merged segment increases the error the least. Figure 2 shows the approximations generated by this strategy. It does an excellent job at identifying the essentially linear portions of the curve and puts the boundaries between component models at the corners. While not shown in the figure, as repeated merges take place, more data is available for each segment. This would allow us to reliably fit submodels more complex than linear segments, such as Bezier curves. It is possible to reliably induce a representation which uses linear segments in some portions and higher order curves in others. Such models potentially have many parameters and would be subject to overfitting if they were learned directly rather than by going through merging steps.

Model merging has an obvious converse in iterative model *splitting*. In the curve example, this top-down approach would start with a single segment and repeatedly split it. This approach sometimes has to make decisions too early and often misses the corners in the curve. Although this is clearly domain-dependent, our experience has been that modelling approaches based on splitting tend to fit the structure of a domain less well than those based on merging.

## 3.2 Model Merging for HMMs

We describe the application of model merging to HMMs in two steps. In the remainder of Section 3 we give a general outline of the algorithm and discuss the theoretical concepts involved. Section 4 goes into the details of the implementation and discusses the various approximations and computational shortcuts used for efficiency.

### 3.2.1 Overview

The model merging method requires three major elements:

1. A method to construct an initial model from data.

2. A way to identify and merge submodels.

3. An error measure to compare the goodness of various candidates for merging, and to limit the generalization.

These elements can be translated to the HMM domain as follows:

1. An initial HMM is constructed as a disjunction of all observed samples. Each sample is represented by dedicated HMM states such that the entire model generates all and only the observed strings.

2. The merging step combines individual HMM states and gives the combined state emission and transition probabilities which are weighted averages of the corresponding distributions for the states which have been merged.

3. The simplest error is the negative logarithm of the model likelihood. Later we show how this is generalized to a Bayesian posterior model probability criterion that provides a principled basis for limiting generalization.

To obtain an initial model from the data, we first construct an HMM which produces exactly the input strings. The start state has as many outgoing transitions as there are strings and each string is represented by a unique path with one state per sample symbol. The probability of entering these paths from the start state is uniformly distributed. Within each path there is a unique transition to the next state, with probability 1. The emission probabilities are 1 for each state to produce the corresponding symbol.

The initial model resulting from this procedure has the property that it assigns each sample a probability equal to its relative frequency, and is therefore a maximum likelihood model for the data, as is generally true for initial models in the model merging methodology. In this sense the initial HMM is also the most *specific* model compatible with the data (modulo weak equivalence among HMMs).

The merging operation, repeatedly applied to pairs of HMM states, preserves the ability to generate all the samples accounted for by the initial model. However, new, unobserved strings may also be generated by the merged HMM. This in turn means that the probability mass is distributed among a greater number (possibly an infinity) of strings, as opposed to just among the sample strings. The algorithm therefore *generalizes* the sample data.

The drop in the likelihood relative to the training samples is a measure of how much generalization occurs. By trying to minimize the change in likelihood, the algorithm performs repeated *conservative* generalizations, until a certain threshold is reached. We will see later that the trade-off between model likelihood and generalization can be recast in Bayesian terms, replacing the simple likelihood thresholding scheme by the maximization of posterior model probability.

### 3.2.2 An example

Consider the regular language $(ab)^+$ and two samples drawn from it, the strings $ab$ and $abab$. Using the above procedure, the algorithm constructs the initial model $M_0$ depicted in Figure 3.

From this starting point, we can perform two merging steps without incurring a drop in the model likelihood.[3] First, states 1 and 3 are merged ($M_1$), followed by 2 and 4 ($M_2$).

Merging two states entails the following changes to a model:

- The old states are removed and the new 'merged' state is added. The old states are called the 'parent' states.
- The transitions from and to the old states are redirected to the new state. The transition probabilities are adjusted to maximize the likelihood.
- The new state is assigned the union of the emissions of the old states and the emission probabilities are adjusted to maximize the likelihood.

In this example we use the convention of numbering the merged state with the smaller of the indices of its parents.

[3] Actually there are two symmetrical sequences of merges with identical result. We have arbitrarily chosen one of them.

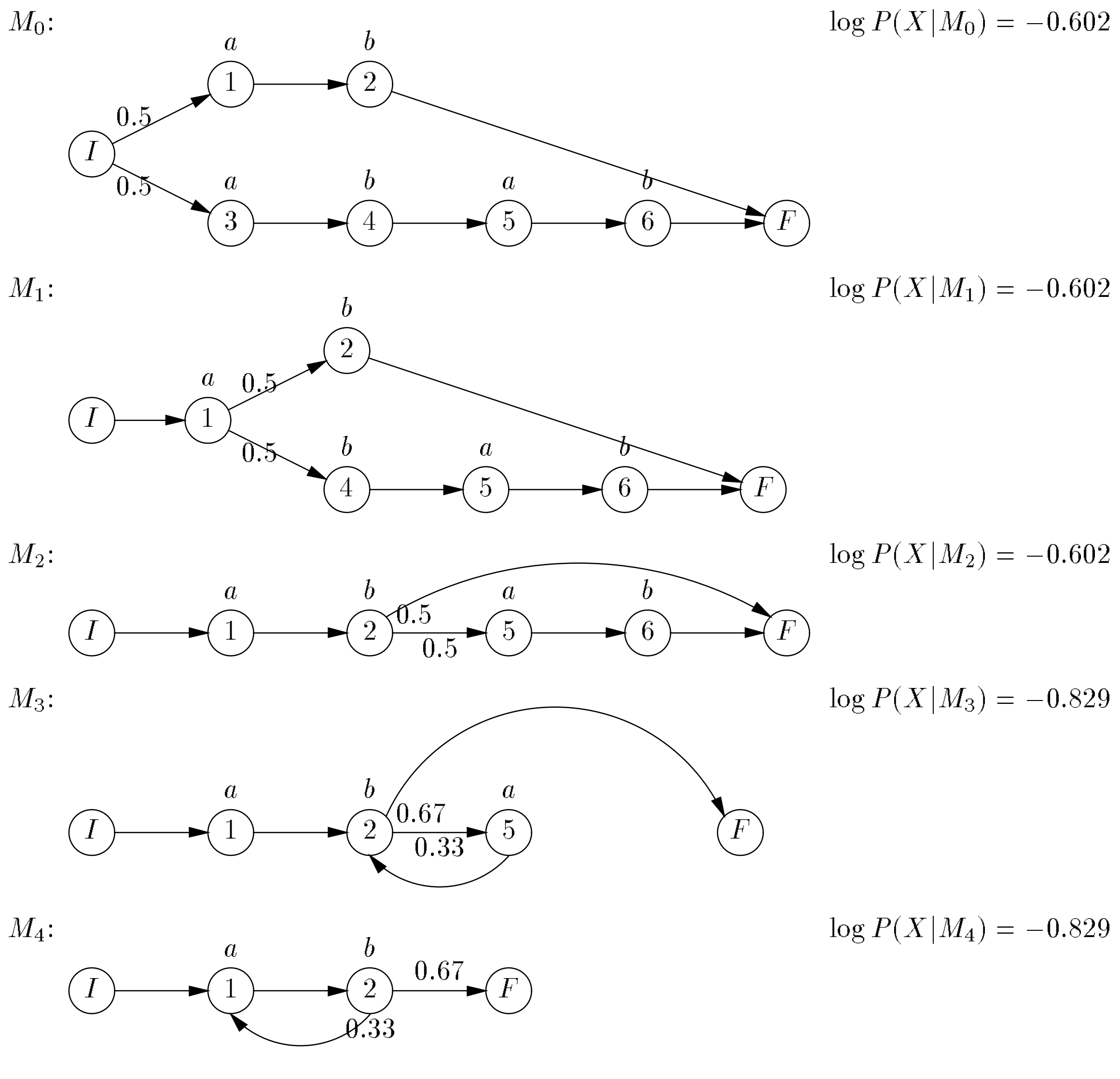

Figure 3: Sequence of models obtained by merging samples $\{ab, abab\}$.

All transitions without special annotations have probability 1. Output symbols appear above their respective states and also carry an implicit probability of 1. For each model, the log likelihood (base 10) is given.

Returning to the example, we now chose to merge states 2 and 6 ($M_3$). This step decreases the log likelihood (from $-0.602$ to $-0.829$) but it is the smallest decrease that can be achieved by any of the potential merges.

Following that, states 1 and 3 can be merged without penalty ($M_4$). The resulting HMM is the minimal model generating the target language $(ab)^+$, but what prevents us from merging further, to obtain an HMM for $\{ab\}^+$ ?

It turns out that merging the remaining two states reduces the likelihood much more drastically than the previous, 'good' generalization step, from $-0.829$ to $-3.465$ (*i.e.*, three decimal orders of magnitude). A preliminary answer, therefore, is to set the threshold small enough to allow only desirable generalizations. A more satisfactory answer is provided by the Bayesian methods described below.

Note that further data may well justify the generalization to a model for $\{ab\}^+$. This *data-driven* character is one of the central aspects of model merging.

A domain-specific justification for model merging in the case of HMMs applies. It can be seen from the example that the structure of the generating HMM can always be recovered by an appropriate sequence of state merges from the initial model, provided that the available data 'covers' all of the generating model, *i.e.*, each emission and transition is exercised at least once. Informally, this is because the initial model is obtained by 'unrolling' the paths used in generating the samples in the target model. The iterative merging process, then, is an attempt to undo the unrolling, tracing a search through the model space back to the generating model. Of course, the best-first heuristic is not guaranteed to find the appropriate sequence of merges, or, less critically, it may result in a model that is only weakly equivalent to the generating model.

### 3.3 Bayesian model merging

Learning from sample data in the model merging framework means generalizing from it. This implies trading off model likelihood against some kind of bias towards 'simpler' models. In the previous formulation simplicity is implicitly defined by the merging operator, *i.e.*, merged models are by definition simpler than unmerged models.

Alternatively, we can express a preference among alternative models (for 'simplicity' or otherwise) in probabilistic terms using the Bayesian notion of *prior probability.* We assume that there exists a distribution $P(M)$ independent of the data that assigns each model $M$ an *a priori* (before the data) probability, which can be understood as expressing a bias. Given some data $X$ we can then endeavor to find the model $M_{\mathrm{MAP}}$ that maximizes the *posterior probability* $P(M|X)$. Bayes' Law expresses the posterior as

$$P(M|X) = \frac{P(M)P(X|M)}{P(X)} \tag{5}$$

Since the data $X$ is fixed, $M_{\mathrm{MAP}}$ maximizes $P(M)P(X|M)$ , where $P(X|M)$ is the familiar likelihood.

This form of Bayesian model inference is therefore a generalization of the Maximum-Likelihood (ML) estimation method, adding the prior $P(M)$ to the expression being maximized. The difference is crucial, as we have seen, since in the domain of structural HMM inference (as opposed to fixed-set parameter estimation) it is always trivially possible to construct a ML model that is uninteresting in that it expresses no generalization.

It is easy to modify the model merging strategy to accommodate the prior probability. Instead of dealing with the model likelihood $P(X|M)$ alone, which is guaranteed to have a maximum at the initial model, we can maximize the posterior probability $P(M|X)$ through merging. Relative to the first formulation, this implies two modifications:

- Each merging step should maximize the posterior $P(M|X)$.
- Merging will continue as long as the posterior keeps *increasing* (as opposed to passing some fixed threshold).

## 3.4 Priors for Hidden Markov Models

From the previous discussion it is clear that the choice of the prior distribution is important since it is the term in (5) that drives generalization. We take the approach that priors should be subject to experimentation and empirical comparison of their ability to lead to useful generalization. The choice of a prior represents an intermediate level of probabilistic modeling, between the global choice of model formalism (HMMs, in our case) and the choice of a particular instance from a model class (*e.g.*, a specific HMM structure and parameters). The model merging approach ideally replaces the usually poorly constrained choice of low-level parameters with a more robust choice of (few) prior parameters. As long as it doesn't assign zero probability to the correct model, the choice of prior is eventually overwhelmed by a sufficient amount of data. In practice, the ability to *find* the correct model may be limited by the search strategy used, in our case, the merging process.

HMMs are a special kind of parameterized graph structure. Unsurprisingly, many aspects of the priors discussed in this section can be found in Bayesian approaches to the induction of graph-based models in other domains (*e.g.*, Bayesian networks (Cooper & Herskovits 1992; Buntine 1991) and decision trees (Buntine 1992)).

### 3.4.1 Structural vs. parameter priors

An HMM can be described in two stages:

1. A model *structure* or *topology* is specified as a set of states, transitions and emissions. Transitions and emissions represent discrete choices as to which paths and outputs can have non-zero probability in the HMM.
2. Conditional on a given structure, the model's continuous probability parameters are specified.

We will write $M = (M_S, \theta_M)$ to describe the decomposition of model $M$ into the structure part $M_S$ and the parameter part $\theta_M$. The model prior $P(M)$ can therefore be written as

$$P(M) = P(M_S)P(\theta_M|M_S)$$

Even this framework leaves some room for choice: one may choose to make the structure specification very general, *e.g.*, by allowing transitions between any two states. The presence of a transition is then given only in the parameter part of the specification as a non-zero probability.

Our approach is to compose a prior for both the structure and the parameters of the HMM as a product of independent priors for each transition and emission multinomial, possibly along with a global factor. Although the implicit independence assumption about the parameters of different states is clearly a simplification, it shouldn't introduce any systematic bias toward any particular model structure. It does, however, greatly simplify the computation and updating of the global posteriors for various model variants, as detailed in Section 4.

The global prior for a model $M$ thus becomes a product

$$P(M) = P(M_G) \prod_{q \in \mathcal{Q}} P(M_S^{(q)}|M_G) P(\theta_M^{(q)}|M_G, M_S^{(q)}) \tag{6}$$

where $P(M_G)$ is a prior for global aspects of the model structure (including, *e.g.*, the number of states), $P(M_S^{(q)})$ is a prior contribution for the structure associated with state $q$, and $P(\theta_M^{(q)}|M_S^{(q)})$ is a prior on the parameters (transition and emission probabilities) associated with state $q$.

Unless otherwise noted, the global factor $P(M_G)$ is assumed to be unbiased, and therefore ignored in the maximization.

### 3.4.2 Parameter priors

The parameters in an HMM with discrete outputs can be described entirely as the parameters for a collection of multinomial distributions. Each transition represents a discrete, finite probabilistic choice of the next state, as do the emissions which choose among output symbols. Let $n$ be the number of choices in a multinomial and, $\boldsymbol{\theta} = (\theta_1, \ldots, \theta_n)$ the probability parameters associated with each choice (only $n-1$ of these parameters are free since $\sum_i \theta_i = 1$).

A standard prior for multinomials is the *Dirichlet distribution*

$$P(\boldsymbol{\theta}) = \frac{1}{B(\alpha_1, \ldots, \alpha_n)} \prod_{i=1}^{n} \theta_i^{\alpha_i - 1} \quad , \tag{7}$$

where $\alpha_1, \ldots, \alpha_n$ are parameters of the prior which can be given an intuitive interpretation (see below). The normalizing constant $B(\alpha_1, \ldots, \alpha_n)$ is the $n$-dimensional Beta function,

$$B(\alpha_1, \ldots, \alpha_n) = \frac{\Gamma(\alpha_1) \cdots \Gamma(\alpha_n)}{\Gamma(\alpha_1 + \cdots + \alpha_n)} \quad .$$

The *prior weights* $\alpha_i$ determine the bias embodied in the prior: the prior expectation of $\theta_i$ is $\frac{\alpha_i}{\alpha_0}$, where $\alpha_0 = \sum_i \alpha_i$ is the *total prior weight.*

One important reason for the use of the Dirichlet prior in the case of multinomial parameters (Cheeseman *et al.* 1988; Cooper & Herskovits 1992; Buntine 1992) is its mathematical expediency. It is a *conjugate prior*, *i.e.*, of the same functional form as the likelihood function for the multinomial. The likelihood for a sample from the multinomial with total observed outcomes $c_1, \ldots, c_n$ is given by[4]

$$P(c_1, \ldots, c_n | \boldsymbol{\theta}) = \prod_{i=1}^{n} \theta_i^{c_i} \quad . \tag{8}$$

This means that the prior (7) and the likelihood (8) combine according to Bayes' law to give an expression for the posterior density that is again of the same form, namely:

$$P(\boldsymbol{\theta} | c_1, \ldots, c_n) = \frac{1}{B(c_1 + \alpha_1, \ldots, c_n + \alpha_n)} \prod_{i=1}^{n} \theta_i^{c_i + \alpha_i - 1} \quad . \tag{9}$$

Furthermore, it is convenient that the integral over the product of (7) and (8) has a closed-form solution.

$$\begin{aligned} \int_{\boldsymbol{\theta}} P(\boldsymbol{\theta}) P(c_1, \ldots, c_n | \boldsymbol{\theta}) d\boldsymbol{\theta} &= \frac{1}{B(\alpha_1, \ldots, \alpha_n)} \int_{\boldsymbol{\theta}} \prod_{i=1}^{n} \theta_i^{c_i + \alpha_i - 1} d\boldsymbol{\theta} \\ &= \frac{B(c_1 + \alpha_1, \ldots, c_n + \alpha_n)}{B(\alpha_1, \ldots, \alpha_n)} \end{aligned} \tag{10}$$

This integral will be used to compute the posterior for a given model structure, as detailed in Sections 3.4.4 and 4.3.

To get an intuition for the effect of the Dirichlet prior it is helpful to look at the two-dimensional case. For $n = 2$ there is only one free parameter, say $\theta_1 = p$, which we can identify with the probability of heads in a biased coin flip ($\theta_2 = 1 - p$ being the probability of tails). Assume there is no *a priori* reason to prefer either outcome, *i.e.*, the prior distribution should be symmetrical about the value $p = 0.5$. This symmetry entails a choice of $\alpha_i$'s which are equal, in our case $\alpha_1 = \alpha_2 = \alpha$. The resulting prior distribution is depicted in Figure 4(a), for various values of $\alpha$. For $\alpha_i > 1$ the prior has the effect of adding $\alpha_i - 1$ 'virtual' samples to the likelihood expression, resulting in a MAP estimate of

$$\hat{\theta}_i = \frac{c_i + \alpha_i - 1}{\sum_j (c_i + \alpha_i - 1)} \quad .$$

For $0 < \alpha_i < 1$ the MAP estimate is biased towards the extremes of the parameter space, $\theta_i = 0$ and $\theta_i = 1$. For $\alpha_i = 1$ the prior is uniform and the MAP estimate is identical to the ML estimate.

Figure 4(b) shows the effect of varying amounts of data $N$ (total number of samples) on the posterior distribution. With no data ($N = 0$) the posterior is identical to the prior, illustrated here for $\alpha = 2$. As $N$ increases the posterior peaks around the ML parameter setting.

**Narrow parameter priors** A natural application of the Dirichlet prior is as a prior distribution over each set of multinomial parameters within a given HMM structure $M_S$.

[4] For unordered samples the expression has to be scaled by the multinomial coefficient $\frac{(c_1 + \cdots + c_n)!}{c_1! \cdots c_n!}$. Because the expression for ordered samples is simpler and constant factors don't make any difference for model comparison we generally use the expression for the simpler, ordered case.

a)

b)

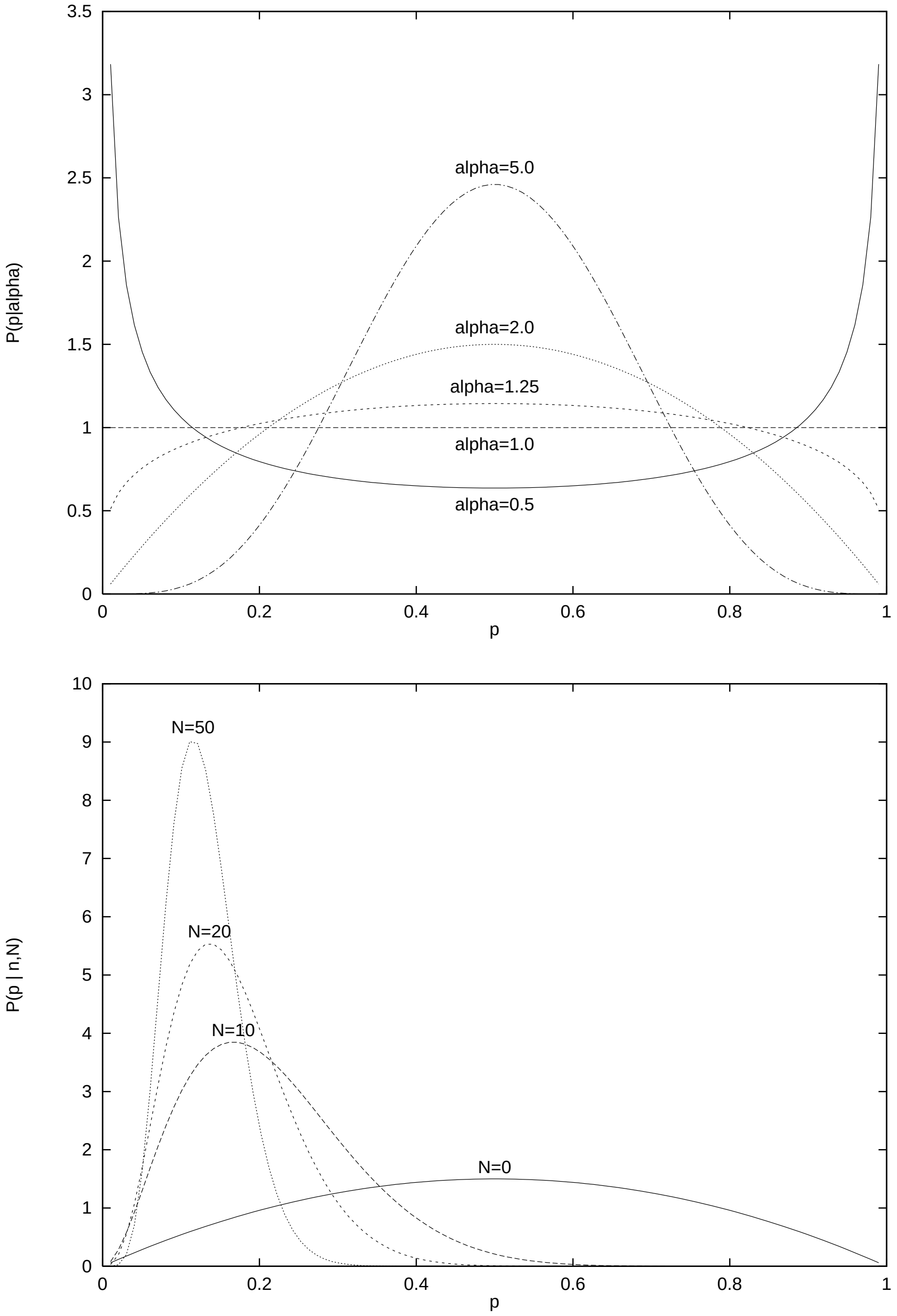

Figure 4: The two-dimensional symmetrical Dirichlet prior.

(a) Prior distributions for various prior weights $\alpha$. (b) Posterior distributions for $\alpha = 2.0$ and various amounts of data $N = c_1 + c_2$ in the proportion $c_1/N = 0.1$.

Relative to equation (6), the parameters of a state $q$ with $n_t^{(q)}$ transitions and $n_e^{(q)}$ emissions contribute a factor

$$P(\theta_M^{(q)}|M_G, M_S^{(q)}) = \frac{1}{B(\alpha_t, \ldots, \alpha_t)} \prod_{i=1}^{n_t^{(q)}} \theta_{qi}^{\alpha_t - 1} \frac{1}{B(\alpha_e, \ldots, \alpha_e)} \prod_{i=1}^{n_e^{(q)}} \theta_{qj}^{\alpha_e - 1} \quad . \tag{11}$$

Here $\theta_{qi}$ are the transition probabilities at state $q$, $i$ ranging over the states that can follow $q$; $\theta_{qj}$ are the emission probabilities in state $q$, $j$ ranging over the outputs emitted by $q$. $\alpha_t$ and $\alpha_e$ are the prior weights for transitions and emissions, respectively, and can be chosen to introduce more or less bias towards a uniform assignment of the parameters.

**Broad parameter priors** In the preceding version the parameters were constrained by the choice of a model structure $M_S$. As indicated earlier, one may instead let the parameters range over all potential transitions (all states in the model) and emissions (all elements of the output alphabet). Dirichlet priors as in equation (11) can still be used, using $n_t^{(q)} = |\mathcal{Q}|$ and $n_e^{(q)} = |\Sigma|$ for all states $q$.

One interesting aspect of this approach is that at least the emission prior weights can be chosen to be non-symmetrical, with prior means

$$E[\theta_i] = \frac{\alpha_i}{\sum_j \alpha_j}$$

adjusted so as to match the empirical fraction of symbol occurrences in the data. This 'empirical Bayes' approach is similar to the setting of prior class probability means in Buntine (1992).

We are already working on the assumption that transitions and emissions are *a priori* independent of each other. It is therefore also possible to use any combination of broad and narrow parameter priors.

### 3.4.3 Structure priors

In the case of broad parameter priors the choice of transitions and emissions is already subsumed by the choice of parameters. The only structural component left open in this case is the number of states $|\mathcal{Q}|$. For example, one might add an explicit bias towards a smaller number of states by setting

$$P(M_S) \propto C^{-|\mathcal{Q}|}$$

for some constant $C > 1$. However, as we will see below, the state-based priors by themselves produce a tendency towards reducing the number of states as a result of Bayesian 'Occam factors' (Gull 1988).

In the case of narrow parameter priors we need to specify how the prior probability mass is distributed among all possible model topologies with a given number of states. For practical reasons it is desirable to have a specification that can be described as a product of individual state-based distributions. This leads to the following approach.

As for the transitions, we assume that each state has on average a certain number of outgoing transitions, $n_t$. We don't have a reason to prefer any of the $|\mathcal{Q}|$ possible target states *a priori*, so each potential transition will be assessed a prior probability of existence of $p_t = \frac{n_t}{|Q|}$. Similarly, each possible emission will have a prior probability of $p_e = \frac{n_e}{|\Sigma|}$, where $n_e$ is the prior expected number of emissions per state.

The resulting structural contribution to the prior for a state $q$ becomes

$$P(M_S^{(q)}|M_G) = p_t^{n_t^{(q)}}(1-p_t)^{|Q|-n_t^{(q)}} p_e^{n_e^{(q)}}(1-p_t)^{|\Sigma|-n_e^{(q)}} \quad . \tag{12}$$

As before, $n_t^{(q)}$ represents the number of transitions from state $q$, and $n_e^{(q)}$ the number of its emissions.

**Minimum Description Length** Especially in the domain of discrete structures, it is useful to remember the standard duality between the Bayesian approach and inference by *Minimum Description Length* (Rissanen 1983; Wallace & Freeman 1987).

Briefly, the maximization of

$$P(M, X) = P(M)P(X|M)$$

implicit in Bayesian model inference is equivalent to minimizing

$$-\log P(M, X) = -\log P(M) - \log P(X|M) \quad .$$

This in turn can be interpreted as minimizing the coding or description lengths of the data $X$ together with an underlying coding model $M$. Here $-\log P(M)$ is the optimal encoding length of the model under the prior, whereas the negative log likelihood $-\log P(X|M)$ corresponds to an optimal code for the data using $M$ as a probabilistic model.

The structural prior (12) above corresponds to a HMM coding scheme in which each transition is encoded by $-\log p_t$ bits, and each emission with $-\log p_e$ bits. Potential transitions and emissions that are missing each take up $-\log(1-p_t)$ and $-\log(1-p_e)$ respectively.

**Description Length priors** Conversely, any (prefix-free) coding scheme for models that assigns $M$ a code length $\ell(M)$ can be used to induce a prior distribution over models with

$$P(M) \propto e^{-\ell(M)} \quad .$$

We can take advantage of this fact to design 'natural' priors for many domains. For example, a natural way to encode the transitions and emissions in an HMM is to simply enumerate them. Each transition can be encoded using $\log(|Q| + 1)$ bits, since there are $|Q|$ possible transitions, plus a special 'end' marker which allows us not to encode the missing transitions explicitly. The total description length for all transitions from state $q$ is thus $n_t^{(q)} \log(|Q|+1)$. Similarly, all emissions from $q$ can be coded using $n_e^{(q)} \log(|\Sigma| + 1)$ bits.[5]

[5]The basic idea of encoding transitions and emissions by enumeration has various more sophisticated variants. For example, one could base the enumeration of transitions on a canonical ordering of states, such that only $\log(n+1) + \log n + \cdots + log(n - n_t + 1)$ bits are required. Or one could use the $k$-out-of-$n$-bit integer coding scheme described in Cover & Thomas (1991) and used for MDL inference in Quinlan & Rivest (1989). Any reasonable Bayesian inference procedure should not be sensitive to such minor difference in the prior, unless it is used with too little data. Our goal here is simply to suggest priors that have reasonable qualitative properties, and are at the same time computationally convenient.

The resulting prior

$$P(M_S^{(q)}|M_G) \propto (|\mathcal{Q}|+1)^{-n_t^{(q)}}(|\Sigma|+1)^{-n_e^{(q)}} \tag{13}$$

has the property that small differences in the number of states matter little compared to differences in the total number of transitions and emissions.

### 3.4.4 Posteriors for HMM structures

The Bayesian approach in its simplest form computes the posterior probability of a fully specified model,

$$P(M|X) \propto P(M)P(X|M) \quad ,$$

and compares alternative models on that basis. If the goal is to find a single model that best represents the data, this approach amounts to a joint maximization of the posterior $P(M|X)$ over both the model structure $M_S$ and its parameters $\theta_M$.

Alternatively, we may want to infer a single HMM $M$ and view it as a representative of a class of HMMs obtained by varying the parameters $\theta_M$ according to their posterior distribution, $P(\theta_M|X, M_S)$. For example, when new data $X'$ arrives, its probability is assessed as a weighted average over all parameter settings.

$$P(X'|X, M_S) = \int_{\theta_M} P(\theta_M|X, M_S)P(X'|M_S, \theta_M)d\theta_M \tag{14}$$

This is motivated by the Bayes-optimal solution to the transductive inference $P(X'|X)$, which would consist of summing not only over all possible parameter settings, but over all possible model structures as well:

$$P(X'|X) = \sum_{M_S} P(M_S|X) \int_{\theta_M} P(\theta_M|X, M_S)P(X'|M_S, \theta_M)d\theta_M \tag{15}$$

Choosing a single structure is an approximation to the full Bayesian solution, *i.e.*, averaging only over a part of the full model space. To optimize this approximation we should choose the model structure $M_S$ which maximizes the associated posterior weight in equation (15),

$$P(M_S|X) \int_{\theta_M} P(\theta_M|X, M_S)d\theta_M = P(M_S|X) \quad .$$

This reasoning suggests changing the objective from maximizing the joint posterior probability of the structure and parameters together, to maximizing the posterior probability of the model structure alone. The desired quantity is obtained by integrating out the 'nuisance' variable $\theta_M$:

$$\begin{aligned} P(M_S|X) &= \frac{P(M_S)P(X|M_S)}{P(X)} \\ &= \frac{P(M_S)}{P(X)} \int_{\theta_M} P(X, \theta_M|M_S)d\theta_M \\ &= \frac{P(M_S)}{P(X)} \int_{\theta_M} P(\theta_M|M_S)P(X|M_S, \theta_M)d\theta_M \quad . \end{aligned} \tag{16}$$

Unfortunately, there seems to be no way to compute this integral exactly in closed form, since $P(X|M_S, \theta_M)$ is a sum over all possible paths through the HMM that can generate $X$, and whose respective probabilities vary with $\theta_M$. In Section 4 we give a solution that relies on the approximation of sample likelihoods by Viterbi paths.

### 3.5 Why are smaller HMMs preferred?

Intuitively, we want an HMM induction algorithm to prefer 'smaller' models over 'larger' ones, other things being equal. This can be interpreted as a special case of 'Occam's razor,' or the scientific maxim that simpler explanations are to be preferred unless more complex explanations are required to explain the data.

Once the notions of model size (or explanation complexity) and goodness of explanation are quantified, this principle can be modified to include a trade-off between the criteria of simplicity and data fit. This is precisely what the Bayesian approach does, since in optimizing the product $P(M)P(X|M)$ a compromise between simplicity (embodied in the prior) and fit to the data (high model likelihood) is found.

But how is it that the HMM priors discussed in the previous section lead to a preference for 'smaller' or 'simpler' models? Two answers present themselves: one has to do with the general phenomenon of 'Occam factors' found in Bayesian inference; the other is related, but specific to the way HMMs partition data for purposes of 'explaining' it. We will discuss each in turn.

#### 3.5.1 Occam factors

Consider the following scenario. Two pundits, $M_1$ and $M_2$, are asked for their predictions regarding an upcoming election involving a number of candidates. Each pundit has his/her own 'model' of the political process. We will identify these models with their respective proponents, and try to evaluate each according to Bayesian principles. $M_1$ predicts that only three candidates, $A$, $B$, and $C$ have a chance to win, each with probability $\theta_{A_1} = \theta_{B_1} = \theta_{C_1} = \frac{1}{3}$. $M_2$ on the hand gives only $A$ and $B$ a realistic chance, with probability $\theta_{A_2} = \theta_{B_2} = \frac{1}{2}$. Candidate $B$ turns out to be the winner. What is the posterior credibility of each pundit?

We marginalize over the (discrete) parameter space of each pundit's predictions. The 'data' $X$ is the outcome of $B$'s winning.

$$
\begin{aligned}
P(M_1|X) &\propto P(M_1)[P(A|M_1)P(X|A) + P(B|M_1)P(X|B) + P(C|M_1)P(X|C)] \\
&= P(M_1)[\theta_{A_1} \cdot 0 + \theta_{B_1} \cdot 1 + \theta_{C_1} \cdot 0] \\
&= P(M_1)\frac{1}{3} \\
P(M_2|X) &\propto P(M_2)[P(A|M_2)P(X|A) + P(B|M_2)P(X|B)] \\
&= P(M_2)[\theta_{A_2} \cdot 0 + \theta_{B_2} \cdot 1] \\
&= P(M_2)\frac{1}{2}
\end{aligned}
$$

Assuming that there is no *a priori* preference, $P(M_1) = P(M_2)$, we conclude that $M_2$ is more likely *a posteriori*. This result, of course, just confirms our intuition that a prophet whose predictions are specific (and true) is more credible than one whose predictions are more general.

The ratio between the allowable range of a model's parameters *a posterior* and *a priori* is known as the *Occam factor* (Gull 1988). In the discrete case these ranges are just the respective numbers of possible parameter settings: $\frac{1}{3}$ versus $\frac{1}{2}$ in the example. For continuous model parameters, the Occam factor penalizes those models in which the parameters have a larger range, or where the parameter space has a higher dimensionality. (This is how the Bayesian approach avoids always picking the model with the largest number of free parameters, which leads to overfitting the data.)

### 3.5.2 Effective amount of data per state

Prior to implementing (an approximation to) the full computation of the structure posterior for HMMs as dictated by equation (16), we had been experimenting with a rather crude heuristic that simply compared the likelihoods for alternative models, but evaluated *at the MAP parameter settings*. The prior used was a Dirichlet of the broad type discussed in Section 3.4.2. As a result, we were entirely neglecting the structural prior itself which favors smaller configurations. Surprisingly, we found this alone to produce a preference for smaller models!

The intuitive reason for this is a combination of two phenomena, one of which is particular to HMMs. As is true in general, the MAP point migrates towards the maximum likelihood setting as the amount of data increases (Figure 4(b)). But in the case of HMMs, the effective amount of data *per state* increases as states are merged! In other words, as the number of states in an HMM shrinks, but total amount of data remains constant, each state will get to 'see' more of the data, on average. Therefore, merging the right states will cause some states to have more data available to them, so as to move the likelihood closer to its maximum value.

## 3.6 The algorithm

After choosing a set of priors and prior parameters, it is conceptually straightforward to modify the simple likelihood-based algorithm presented in Section 3.2 to accommodate the Bayesian approach discussed above. The best-first HMM merging algorithm takes on the following generic form.

#### Best-first merging (batch version)

A. Build the initial, maximum-likelihood model $M_0$ from the dataset $X$.

B. Let $i := 0$. Loop:

1. Compute a set of candidate merges $K$ among the states of model $M_i$.
2. For each candidate $k \in K$ compute the merged model $k(M_i)$, and its posterior probability $P(k(M_i)|X)$.
3. Let $k^*$ be the merge that maximizes $P(k(M_i)|X)$. Then let $M_{i+1} := k^*(M_i)$.
4. If $P(M_{i+1}|X) < P(M_i|X)$, break from the loop.

5. Let $i := i + 1$.

C. Return $M_i$ as the induced model.

In this formulation 'model' can stand for either 'model structure + parameters' or, as suggested above, just 'model structure.' In discussing our implementation and results we assume the latter unless explicitly stated otherwise.

Note that many of the computational details are not fleshed out here. Important implementation strategies are described in Section 4.

The number of potential merges in step B.2, $|K|$, is the biggest factor in the total amount of computation performed by the algorithm. Although $|K|$ can sometimes be reduced by domain-specific constraints (Section 4.2), it is generally $O(|Q|^2)$. Because $|Q|$ grows linearly with the total length of the samples, this version of the merging algorithm is only feasible for small amounts of data.

An alternative approach is to process samples incrementally, and start merging after a small amount of new data has been incorporated. This keeps the number of states, and therefore the number of candidates, small. If learning is successful, the model will stop growing eventually and reach a configuration that accounts for almost all new samples, at which point no new merges are required. (Figure 17 shows a size profile during incremental merging in one of our applications.) The incremental character is also more appropriate in scenarios where data is inherently incomplete and an on-line learning algorithm is needed that continuously updates a working model.

**Best-first merging (on-line version)**
Let $M_0$ be the empty model. Let $i := 0$. Loop:

A. Get some new samples $X_i$ and incorporate into the current model $M_i$.

B. Loop:

1. Compute a set of candidate merges $K$ from among the states of model $M_i$.
2. For each candidate $k \in K$ compute the merged model $k(M_i)$ and its posterior probability $P(k(M_i)|X)$.
3. Let $k^*$ be the merge that maximizes $P(k(M_i)|X)$. Then let $M_{i+1} := k^*(M_i)$.
4. If $P(M_{i+1}|X) < P(M_i|X)$, break from the loop.
5. Let $i := i + 1$.

C. If the data is exhausted, break from the loop and return $M_i$ as the induced model.

Incremental merging might in principle produce results worse than the batch version since the evaluation step doesn't have as much data at its disposal. However, we didn't find this to be a significant disadvantage in practice. One can optimize the number of samples incorporated in each step A (the *batch size*) for overall speed. This requires balancing the gains due to smaller model size against the constant overhead of each execution of step B. The best value will depend on the data and how much merging is actually possible on each iteration; we found between 1 and 10 samples at a time to be good choices.

One has to be careful not to start merging with extremely small models, such as that resulting from incorporating only a few short samples. Many of the priors discussed earlier contain logarithmic terms that approach singularities ($\log 0$) in this case, which can produce poor results, usually by leading to extreme merging. That can easily be prevented by incorporating a larger number of samples (say, 10 to 20) before going on to the first merging step.

Further modifications to the simple best-first search strategy are discussed in Section 4.5.

# 4 Implementation Issues

In this section we elaborate on the implementation of the various steps in the generic HMM merging algorithm presented in Section 3.6.

## 4.1 Efficient sample incorporation

In the simplest case this step creates a dedicated state for each instance of a symbol in any of the samples in $X$. These states are chained with transitions of probability 1, such that a sample $x_1 \ldots x_\ell$ is generated by a state sequence $q_1, \ldots, q_\ell$. $q_1$ can be reached from $q_I$ via a transition of probability $\frac{1}{|X|}$, where $|X|$ is the total number of samples. State $q_\ell$ connects to $q_\ell$ with probability 1. All states $q_i$ emit their corresponding output symbol $x_i$ with probability 1.

In this way, repeated samples lead to multiple paths through the model, all generating the sample string. The total probability of a string $x$ according to the initial model is thus $\frac{c(x)}{|X|}$, *i.e.*, the relative frequency of string $x$. It follows that the initial model constitutes a maximum likelihood model for the data $X$.

Note that corresponding states in equivalent paths can be merged without loss of model likelihood. This is generally what the merging loop does in its initial passes.

A trivial optimization at this stage is to avoid the initial multiplicity of paths and check for each new sample whether it is already accounted for by an existing path. If so, only the first transition probability has to be updated.

The idea of shortcutting the merging of samples into the existing model could be pursued further along the lines of Thomason & Granum (1986). Using an extension of the Viterbi algorithm, the new sample can be aligned with the existing model states, recruiting new states only where necessary. Such an alignment couldn't effect all the possible merges, *e.g.*, it wouldn't be able to generate loops, but it could further reduce the initial number of states in the model, thereby saving computation in subsequent steps.[6]

## 4.2 Computing candidate merges

The general case here is to examine all of the $\frac{1}{2}|\mathcal{Q}|(|\mathcal{Q}| - 1)$ possible pairs of states in the current model. The quadratic cost in the number of states explains the importance of the various strategies to keep the number of states in the initial model small.

We have explored various application specific strategies to narrow down the set of worth-while candidates. For example, if the cost of a merge is usually dominated by the cost of merging the output distributions of the states involved, we might index states according to their emission characteristics and consider only pairs of states with similar outputs. This constraint can be removed later after all other merging possibilities have been exhausted. The resulting strategy (first merging only same-outputs states, followed by general merging) not only speeds up the algorithm, it is also generally a good heuristic in incremental merging to prevent premature merges that are likely to be assessed differently in the light of new data.

[6]This optimization is as yet unimplemented.

Sometimes hard knowledge about the target model structure can further constrain the search. For example, word models for speech recognition are usually not allowed to generate arbitrary repetitions of subsequences (see Section 6.2). All merges creating loops (perhaps excepting self-loops) can therefore be eliminated in this case.

### 4.3 Model evaluation using Viterbi paths

To find the posterior probability of a potential model, we need to evaluate the structural prior $P(M_S)$ and, depending on the goal of the maximization, either find the maximum posterior probability (MAP) estimates for the model parameters $\theta_M$, or evaluate the integral for $P(X|M_S)$ given by equation (16).

MAP estimation of HMM parameters could be done using the Baum-Welch iterative reestimation (EM) method, by taking the Dirichlet prior into account in the reestimation step. However, this would would require an EM iteration for each candidate model, taking time proportional to the number all samples incorporated into the model.

Evaluation of $P(X|M_S)$, on the other hand, has no obvious exact solution at all, as discussed in Section 3.4.4.

In both cases the problem is greatly simplified if we use the Viterbi approximation, *i.e.*, the assumption that the probability of any given sample is due primarily to a single generation path in the HMM (Section 2.5).

**Likelihood computation** The exact model likelihood relative to a dataset $X$ is given by the product of the individual sample probabilities, each of which is given by equation (1).

$$\begin{aligned} P(X|M) &= \prod_{x \in X} P(x|M) \\ &= \prod_{x \in X} \sum_{q_1 \ldots q_\ell \in Q^\ell} p(q_I \to q_1) p(q_1 \uparrow x_1) p(q_1 \to q_2) \ldots p(q_\ell \uparrow x_\ell) p(q_\ell \to q_F) \end{aligned}$$

where $\ell$ is the length of sample $x$ and $q_1 \ldots q_\ell$ denotes a path through the HMM, given as a state sequence.

The Viterbi approximation implies replacing the inner summations by the terms with the largest contribution:

$$P(X|M) \approx \prod_{x \in X} \max_{q_1 \ldots q_\ell \in Q^\ell} p(q_I \to q_1) p(q_1 \uparrow x_1) p(q_1 \to q_2) \ldots p(q_\ell \uparrow x_\ell) p(q_\ell \to q_F)$$

The terms in this expression can be conveniently grouped by states, leading to the form

$$P(X|M) \approx \prod_{q \in Q} \left( \prod_{q' \in Q} p(q \to q')^{c(q \to q')} \prod_{\sigma \in \Sigma} p(q \uparrow \sigma)^{c(q \uparrow \sigma)} \right) \tag{17}$$

where $c(q \to q')$ and $c(q \uparrow \sigma)$ are the total counts of transitions and emissions occurring along the Viterbi paths associated with the samples in $X$. We use the notation $c^{(q)}$ for the collection of Viterbi counts associated with state $q$, so the above can be expressed more concisely as

$$P(X|M) = \prod_{q \in Q} P(c^{(q)}|M)$$

**MAP parameter estimation** To estimate approximate MAP parameter settings based on Viterbi path counts, the maximum-likelihood estimates as given by (2) and (3) are modified to include the 'virtual' samples provided by the Dirichlet priors:

$$\hat{p}(q \to q') = \frac{c(q \to q') + \alpha_t - 1}{\sum_{s \in Q}[c(q \to s) + \alpha_t - 1]} \tag{18}$$

$$\hat{p}(q \uparrow \sigma) = \frac{c(q \uparrow \sigma) + \alpha_e - 1}{\sum_{\rho \in \Sigma}[c(q \uparrow \rho) + \alpha_e - 1]}. \tag{19}$$

The $\alpha$'s are the prior proportions associated with the Dirichlet distributions for transitions and emissions respectively, as given in equation (11). (These are here assumed to be uniform for simplicity, but need not be.)

Note that the summations in (18) and (19) are over the entire set of possible transitions and emissions, which corresponds to a broad parameter prior. These summations have to be restricted to the transitions and emissions in the current model structure for narrow parameter priors.

**Structure posterior evaluation** To implement model comparison based on the posterior probabilities of the HMM structures (Section 3.4.4) we need to approximate the integral

$$P(X|M_S) = \int_{\theta_M} P(\theta_M|M_S) P(X|M_S, \theta_M) d\theta_M$$

We will apply the usual Viterbi approximation to $P(X|M_S, \theta_M)$, and assume in addition that the Viterbi paths do not change as $\theta_M$ varies. This approximation will be grossly inaccurate for broad parameter priors, but seems reasonable for narrow priors, where the topology largely determines the Viterbi path. More importantly, we expect this approximation to introduce a systematic error that does not bias the evaluation metric for or against any particular model structure, especially since the models being compared have only small structural differences.

The Viterbi-based integral approximation can now be written as

$$P(X|M_S) \approx \int_{\theta_M} P(\theta_M|M_S) \prod_{x \in Q} P(V(x)|M_S, \theta_M) d\theta_M,$$

$V(x)$ being the Viterbi path associated with $x$. The parameters $\theta_M$ can now be split into their parts by state, $\theta_M = (\theta_M^{(q_1)}, \ldots, \theta_M^{(q_N)})$, and the integral rewritten as

$$P(X|M_S) \approx \int_{\theta_M^{(q_1)}} \cdots \int_{\theta_M^{(q_N)}} \left( \prod_{q \in Q} P(\theta_M^{(q)}|M_S) \prod_{q \in Q} P(c^{(q)}|M_S, \theta_M^{(q)}) \right) d\theta_M^{(q_1)} \ldots d\theta_M^{(q_N)}$$

$$= \prod_{q \in Q} \int_{\theta_M^{(q)}} P(\theta_M^{(q)}) P(c^{(q)} | M_S, \theta_M^{(q)}) d\theta_M^{(q)} \tag{20}$$

The integrals in the second expression can be evaluated in closed form by instantiating the generic formula for Dirichlet priors given in (10).

**Optimistic Viterbi path updating** So far, the Viterbi approximation has allowed us to decompose each of the likelihood, estimation, posterior evaluation problems into a form that allows computation by parts organized around states. To take full advantage of this fact we also need a way to update the Viterbi counts $c^{(q)}$ efficiently during merging. In particular, we want to avoid having to reparse all incorporated samples using the merged model. The approach taken here is to update the Viterbi counts associated with each state optimistically, *i.e.*, assuming that merging preserves the Viterbi paths.

During initial model creation the Viterbi counts are initialized to one, corresponding to the one sample that each state was created for. (If initial states are shared among identical samples the initial counts are set to reflect the multiplicity of the samples.) Subsequently, when merging states $q_1$ and $q_2$, the corresponding counts are simply added and recorded as the counts for the new state. For example, given $c(q_1 \to q')$ and $c(q_2 \to q')$ in the current model, the merged state $q_3$ would be assigned a count

$$c(q_3 \to q') = c(q_1 \to q') + c(q_2 \to q') \quad .$$

This is correct if all samples with Viterbi paths through the transitions $q_1 \to q$ and $q_2 \to q'$ retain their most likely paths in the merged model, simply replacing the merged states with $q_3$, and no other samples change their paths to include $q_3 \to q'$.

This path preservation assumption is not strictly true but holds most of the time, since the merges actually chosen are those that collapse states with similar distributions of transition and emission probabilities. The assumption can be easily verified, and the counts corrected, by reparsing the training data from time to time.

In an incremental model building scenario, where new samples are available in large number and incorporated one by one, interleaved with merging, one might not want to store all data seen in the past. In this case an exponentially decaying average of Viterbi counts can be kept instead. This has the effect that incorrect Viterbi counts will eventually fade away, being replaced by up-to-date counts obtained form parsing more recent data with the current model.

**Incremental model evaluation** Using the techniques described in the previous sections, the evaluation of a model variant due to merging is now possible in $O(|Q| + |\Sigma|)$ amortized time, instead of the $O((|Q| + |\Sigma|) \cdot |Q| \cdot |X|)$ using a naive implementation.

Before evaluating specific candidates for merging, we compute the contributions to the posterior probability by each state in the current model. The prior will usually depend on the total number of states of the model; it is set to the current number minus 1 in these computations, thereby accounting for the prospective merge. The total computation of these contributions is proportional to the number of states and transitions, *i.e.* $O((|Q| + |\Sigma|) \cdot |Q|)$. For each potential merge we then determine the parts of the model it affects; these are precisely the transitions and emissions from the merged states, as well as transitions into the merged states. The total number of HMM elements affected is at most $O(|\mathcal{Q}| + |\Sigma|)$. For all priors considered here, as well as the likelihood computations of (17) and (20), the old quantities can be updated by subtracting off the terms corresponding to old model elements and adding in the terms for the merged HMM. (The computation is based on addition rather than multiplication since logarithms are used for simplicity and accuracy).[7]

Since the initial computation of state contributions is shared among all the $O(|\mathcal{Q}|^2)$ potential merges, the amortized time per candidate will also be on the order $|\mathcal{Q}| + |\Sigma|$. Note that this is a worst case cost that is not realized if the HMMs are sparse as is usual. If the number of transitions and emissions on each state is bounded by a constant, the computation will also require only constant time.

### 4.4 Global prior weighting

As explained previously, the merging strategy trades off generalization for fit to the data. Generalization is driven by maximizing the prior contribution, whereas the data is fit by virtue of maximizing the likelihood. In practice it is convenient to have a single parameter which controls the balance between these two factors, and thereby controls when generalization should stop.

From the logarithmic version of Bayes' law (5) we obtain

$$\log P(M) + \log P(X|M)$$

as the quantity to be maximized. To obtain such a global control parameter for generalization we can modify this to include a *prior weight* $\lambda$:

$$\lambda \log P(M) + \log P(X|M) \tag{21}$$

For $\lambda > 1$ the algorithm will stop merging later, and earlier for $\lambda < 1$.

The global prior weight has an intuitive interpretation as the reciprocal of a 'data multiplier.' Since the absolute, constant scale of the expression in (21) is irrelevant to the maximization, we can multiply by $\frac{1}{\lambda}$ to get

$$\log P(M) + \frac{1}{\lambda} \log P(X|M) = \log P(M) + \log P(X|M)^{\frac{1}{\lambda}}$$

This corresponds to the posterior given the data $X$ repeated $\frac{1}{\lambda}$ times. In other words, by lowering the prior weight we pretend to have more data from the same distribution than we actually observed, thereby decreasing the tendency for merging to generalize beyond the data. We will refer to the actual number of samples $|X|$ multipled by $\frac{1}{\lambda}$ as the *effective sample size*. The quantity $\frac{1}{\lambda}$ is thus equivalent to the multiplier $c$ used in Quinlan & Rivest (1989) to model the 'representativeness' of the data.

Global prior weighting is extremely useful in practice. A good value for $\lambda$ can be found by trial and error for a given amount of data, by starting with a small value and increasing

[7] The evaluation of (20) involves computing multidimensional Beta functions, which are given as products of Gamma functions, one for each transition or emission. Therefore the addition/subtraction scheme can be used for incremental computation here as well. In practice this may not be worth the implementation effort if the absolute computational expense is small.

$\lambda$ successively, while cross-validating or inspecting the results. At each stage the result of merging can be used as the initial model for the next stage, thereby avoiding duplication of work.

Besides as a global generalization control, $\lambda$ was also found to be particularly helpful in counteracting one of the potential shortcomings of incremental merging. Since incremental merging has to make decisions based on a subset of the data, it is especially important to prevent overgeneralization during the early stages. We can adjust $\lambda$ depending on the number of samples processed to always maintain a minimum effective sample size during incremental merging, thereby reducing the tendency to overgeneralize based on few samples. This principle implies a gradual increase of $\lambda$ as more samples are incorporated. An application of this is reported in Section 6.1.

### 4.5 Search issues

Section 3.6 has described the two basic best-first search strategies, *i.e.*, batch versus incremental sample processing. Orthogonal to that choice are various extensions to the search method to help overcome local posterior probability maxima in the space of HMM structures constructed by successive merging operations.

By far the most common problem found in practice is that the stopping criterion is triggered too early, since a single merging step alone decreases the posterior model probability, although additional related steps might eventually increase it. This happens although in the vast majority of cases the first step is in the right direction. The straightforward solution to this problem is to add a 'lookahead' to the best-first strategy. The stopping criterion is modified to trigger only after a fixed number of steps $> 1$ have produced no improvement; merging still proceeds along the best-first path. Due to this, the lookahead depth does not entail an exponential increase in computation as a full tree search would. The only additional cost is the work performed by looking ahead in vain at the end of a merging sequence. That cost is amortized over several samples if incremental merging with a batch size $> 1$ is being used.

Best-first merging with lookahead has been our method of choice for almost all applications, using lookaheads between 2 and 5. However, we have also experimented with *beam search* strategies. In these, a *set* of working models is kept at each time, either limited in number (say, the top $K$ scoring ones), or by the difference in score to the current best model. On each inner loop of the search algorithm, all current models are modified according to the possible merges, and among the pool thus generated the best ones according to the beam criterion are retained. (By including the unmerged models in the pool we get the effect of a lookahead.)

Some duplication of work results from the fact that different sequences of merges can lead to the same final HMM structure. To remove such gratuitous duplicates from the beam we attach a list of *disallowed merges* to each model, which is propagated from a model to its successors generated by merging. Multiple successors of the same model have the list extended so that later successors cannot produce identical results from simply permuting the merge sequence.

The resulting beam search version of our algorithm does indeed produce superior results on data that requires aligning long substrings of states, and where the quality of the align-ment can only be evaluated after several coordinated merging steps. On the other hand, beam search is considerably more expensive than best-first search and may not be worth a marginal improvement.

All results in Section 6 were obtained using best-first search with lookahead. Nevertheless, improved search strategies and heuristics for merging remain an important problem for future research.

## 5 Related Work

Many of the ideas used in our approach to Bayesian HMM induction are not new by themselves, and can be found in similar forms in the vast literatures on grammar induction and statistical inference.

At the most basic level we have the concept of state merging, which is implicit in the notion of state equivalence classes, and as such is pervasively used in much of automata theory (Hopcroft & Ullman 1979). It has also been applied to the induction of non-probabilistic automata (Angluin & Smith 1983).

Still in the field of non-probabilistic automata induction, Tomita (1982) has used a simple hill-climbing procedure combined with a goodness measure based on positive/negative samples to search the space of possible models. This strategy is obviously similar in spirit to our best-first search method (which uses a probabilistic goodness criterion based on positive samples alone).

The incremental version of the merging algorithm, in which samples are incorporated into a preliminary model structure one at a time, is similar in spirit (but not in detail) to the automata learning algorithm proposed by Porat & Feldman (1991), which induces finite-state models from positive-only, lexicographically ordered samples.

The Bayesian approach to grammatical inference goes back at least to Horning (1969), where a procedure is proposed for finding the grammar with highest posterior probability given the data, using an enumeration of all candidate models in order of decreasing prior probability. While this procedure can be proven to converge to the maximum posterior probability grammar after a finite number of steps, it was found to be impractical when applied to the induction of context-free grammars. Horning's approach can be applied to any enumerable grammatical domain, but there is no reason to believe that the simple enumerative approach would be feasible in any but the most restricted applications. The HMM merging approach can be seen as an attempt to make the Bayesian strategy workable by operating in a more data-driven manner, while sacrificing optimality of the result.

A probabilistic approach to HMM structure induction that is probably closest to ours is described by Thomason & Granum (1986). The basic idea is to incrementally build a model structure by incorporating new samples using an extended form of Viterbi alignment. New samples are aligned to the existing model so as to maximize their likelihood, while allowing states to be inserted or deleted for alignment purposes. The procedure is limited to HMMs that have a left-to-right ordering of states, however; in particular, no loops are allowed. In a sense this approach can be seen as an approximation to Bayesian HMM merging for this special class of models. The approximation in this case is twofold: the likelihood (not the posterior) is maximized, and only the likelihood of a single sample (rather than the entire data set) is considered.

Haussler *et al.* (1992) apply HMMs trained by the Baum-Welch method to the problem of protein primary structure alignment. Their model structures are mostly of a fixed, linear form, but subject to limited modification by a heuristic that insert states ('stretches' the model) or deletes states ('shrinks' the model) based on the estimated probabilities.

Somewhat surprisingly, the work by Brown *et al.* (1992) on the construction of class-based $n$-gram models for language modeling can also be viewed as a special case of HMM merging. A class-based $n$-gram grammar is easily represented as an HMM, with one state per class. Transition probabilities represent the conditional probabilities between classes, whereas emission probabilities correspond to the word distributions for each class (for $n > 2$, higher-order HMMs are required). The incremental word clustering algorithm given in (Brown *et al.* 1992) then becomes an instance of HMM merging, albeit one that is entirely based on likelihoods.[8]

# 6 Evaluation

We have evaluated the HMM merging algorithm experimentally in a series of applications. Such an evaluation is essential for a number of reasons:

- The simple priors used in our algorithm give it a general direction, but little specific guidance, or may actually be misleading in practical cases, given finite data.
- Even if we grant the appropriateness of the priors (and hence the optimality of the Bayesian inference procedure in its ideal form), the various approximations and simplifications incorporated in our implementation could jeopardize the result.
- Using real problems (and associated data), it has to be shown that HMM merging is a practical method, both in terms of results and regarding computational requirements.

We proceed in three stages. First, simple formal languages and artificially generated training samples are used to provide a proof-of-concept for the approach. Second, we turn to real, albeit abstracted data derived from the TIMIT speech database. Finally, we give a brief account of how HMM merging is embedded in an operational speech understanding system to provide multiple-pronunciation models for word recognition.[9]

## 6.1 Case studies of finite-state language induction

### 6.1.1 Methodology

In the first group of tests we performed with the merging algorithm, the objective was twofold: we wanted to assess empirically the basic soundness of the merging heuristic and the best-first search strategy, as well as to compare its structure finding abilities to the traditional Baum-Welch method.

[8] Furthermore, after becoming aware of their work, we realized that the scheme Brown *et al.* (1992) are using for efficient recomputation of likelihoods after merging is essentially the same as the one we were using for recomputing posteriors (subtracting old terms and adding new ones).

[9] All HMM drawings in this section were produced using an *ad hoc* algorithm that optimizes layout using best-first search based on a heuristic quality metric (no Bayesian principles whatsoever were involved). We apologize for not taking the time to hand-edit some of the more problematic results, but believe the quality to be sufficient for expository purposes.

To this end, we chose a number of relatively simple regular languages, produced stochastic versions of them, generated artificial corpora, and submitted the samples to both induction methods. The probability distribution for a target language was generated by assigning uniform probabilities at all choice points in a given HMM topology.

The induced models were compared using a variety of techniques. A simple quantitative comparison is obtained by computing the log likelihood on a test set. This is proportional to the negative of (the empirical estimate of) the cross-entropy, which reaches a minimum when the two distributions are identical.

To evaluate the HMM topology induced by Baum-Welch training, the resulting models are *pruned*, *i.e.*, transitions and emissions with probability close to zero are deleted. The resulting structure can then be compared with that of the target model, or one generated by merging. The pruning criterion used throughout was that a transition or emission had an expected count of less than $10^{-3}$ given the training set.

Specifically, we would like to check that the resulting model generates exactly the same discrete language as the target model. This can be done empirically (with arbitrarily high accuracy) using a simple Monte-Carlo experiment. First, the target model is used to generate a reasonably large number of samples, which are then parsed using the HMM under evaluation. Samples which cannot be parsed indicate that the induction process has produced a model that is not sufficiently general. This can be interpreted as *overfitting* the training data.

Conversely, we can generate samples from the HMM in question and check that these can be parsed by the target model. If not, the induction has *overgeneralized.*

In some cases we also inspected the resulting HMM structures, mostly to gain an intuition for the possible ways in which things can go wrong. Some examples of this are presented below.

Note that the outcome of the Baum-Welch algorithm may (and indeed does, in our experience) vary greatly with the initial parameter settings. We therefore set the initial transition and emission probabilities randomly from a uniform distribution, and repeated each Baum-Welch experiment 10 times. Merging, on the other hand, is deterministic, so for each group of runs only a single merging experiment was performed for comparison purposes.

Another source of variation in the Baum-Welch method is the fixed total number of model parameters. In our case, the HMMs are fully parameterized for a given number of states and set of possible emissions; so the number of parameters can be simply characterized by the number of states in the HMM.[10]

In the experiments reported here we tried two variants: one in which the number of states was equal to that of the target model (the minimal number of states for the language at hand), and a second one in which additional states were provided.

Finally, the nature of the training samples was also varied. Besides a standard random sample from the target distribution we also experimented with a 'minimal' selection of representative samples, chosen to be characteristic of the HMM topology in question. This selection is heuristic and can be characterized as follows: "From the list of all strings generated by the target model, pick a minimal subset, in order of decreasing probability, such that each transition and emission in the target model is exercised at least once, and such that each looping transition is exemplified by a sample that traverses the loop at least twice." The minimal training samples this rule produces are listed below, and do in fact seem to be intuitively representative of their respective target models.

[10] By convention, we exclude initial and final states in these counts.

### 6.1.2 Priors and merging strategy

A uniform strategy and associated parameter settings were used in all merging experiments.

The straightforward description length prior for HMM topologies from Section 3.4.3, along with a narrow Dirichlet prior for the parameters (Section 3.4.2) were used to drive generalization. The total Dirichlet prior weight $\alpha_0$ for each multinomial was held constant at $\alpha_0 = 1$, which biases the probabilities to be non-uniform (in spite of the target models used). The objective function in the maximization was the posterior of the HMM structure, as discussed in Section 3.4.4.

Merging proceeded using the incremental strategy described in Section 3.6 (with batch size 1), along with several of the techniques discussed earlier. Specifically, incremental merging was constrained to be among states with identical emissions at first, followed by an unconstrained batch merging phase. The global prior weight $\lambda$ was adjusted so as to keep the effective sample size constant at 50. In accordance with the rationale given in Section 4.4, this gradually increases the prior weight during the incremental merging phase, thereby preventing early overgeneralization.

The search strategy was best-first with 5 steps lookahead.

### 6.1.3 Case study I

The first test task was to learn the regular language $ac^*a \cup bc^*b$, generated by the HMM in Figure 5.[11] It turns out that the key difficulty in this case is to find the dependency between first and last symbols, which can be separated by arbitrarily long sequences of intervening (and non-distinguishing) symbols.[12]

The minimal training sample used for this model consisted of 8 strings:

$aa$
$bb$
$aca$
$bcb$
$acca$
$bccb$
$accca$
$bcccb$

[11] We use standard regular expression notation to describe finite-state languages: $x^k$ stands for $k$ repetitions of the string $x$, $x^*$ denotes 0 or more repetitions of $x$, $x^+$ stands for 1 or more repetitions, and $\cup$ is the disjunction (set union) operator.

[12] This test model was inspired by finite-state models with similar characteristics that have been the subject of investigations into human language learning capabilities (Reber 1969; Cleeremans 1991)

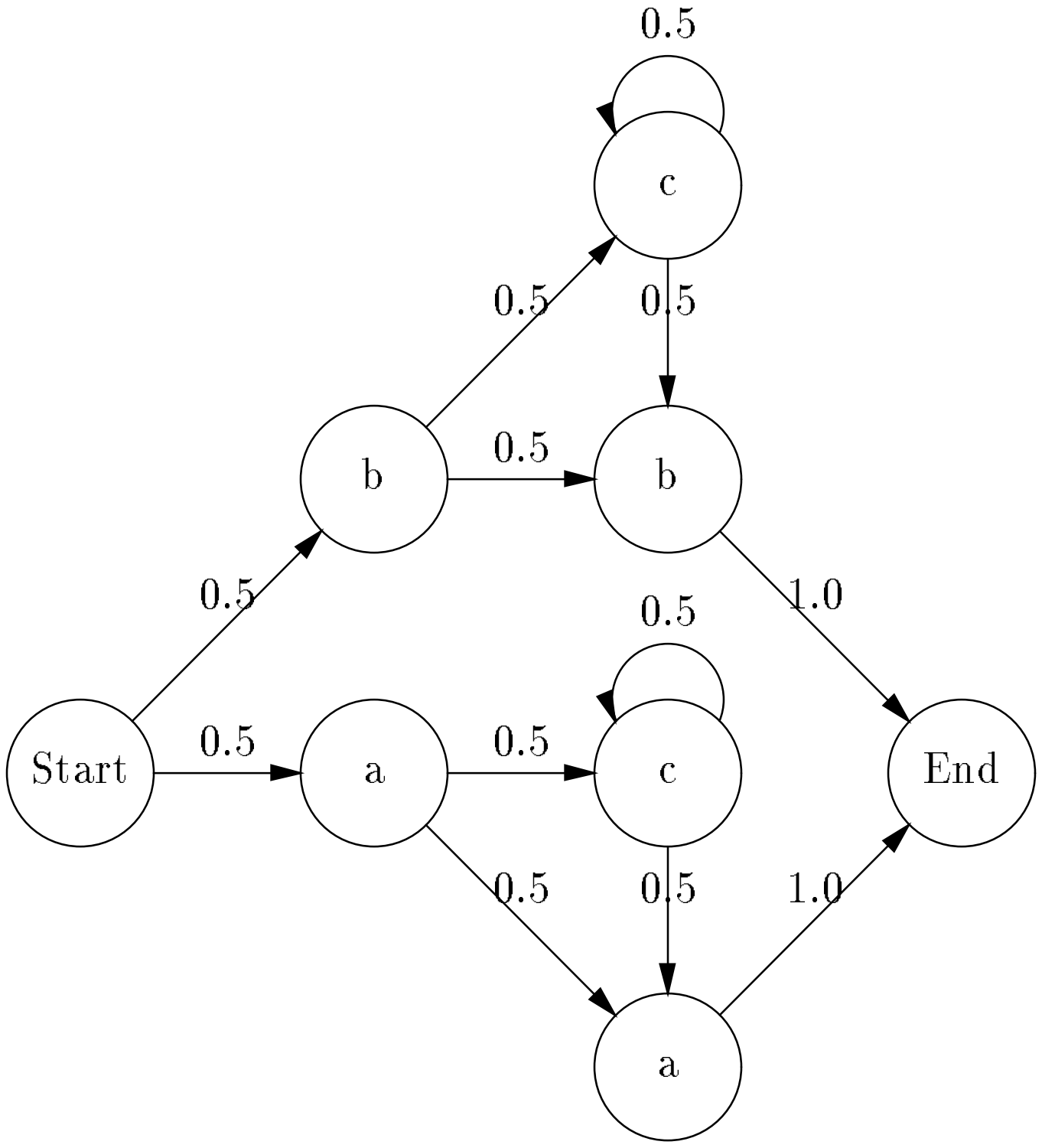

Figure 5: Case study I: HMM generating the test language $ac^*a \cup bc^*b$.

Alternatively, a sample of 20 random strings was used.

The results of both the merging and the Baum-Welch runs are summarized by the series of plots in Figure 6. The plots in the left column refer to the minimal training sample runs, whereas the right column shows the corresponding data for the random 20 string sample runs. Each plot shows a quantitative measure of the induced models' performance, such that the $x$-axis represents the various experiments. In each case, the left-most data point (to the left of the vertical bar) represents the single merging (M) run, followed by 10 data points for repeated Baum-Welch runs with minimal number of states (BW6), and 10 more data points for Baum-Welch runs with 10 states (BW10).

The top row plots the log likelihood on a 100 sample test set; the higher the value, the lower the relative entropy (Kullback-Leibler distance) between the distribution of strings generated by the target model and that embodied by the induced model. To obtain comparable numbers, the probabilities in both the merged model and those estimated by Baum-Welch are set to their ML estimates (ignoring the parameter prior used during merging).

The second row of plots shows the results of parsing the same 100 samples using the discretized induced model topologies, *i.e.*, the number of samples successfully parsed. A score of less than 100 means that the induced model is too specific.

The third row of plots shows the converse parsing experiment: how many out of 100 random samples generated by each induced model can be parsed by the target model. (Note that these 100 samples therefore are not the same across runs.) Therefore, a score of less than 100 indicates that the induced model is overly general.

Note that we use the terms 'general' and 'specific' in a loose sense here which includes cases where two models are not comparable in the set-theoretic sense. In particular, a model can be both 'more general' and 'more specific' than the target model.

When evaluating the structural properties of a model we consider as a 'success' those which neither overgeneralize nor overfit. Such models invariably also have a log likelihood close to optimal. The log likelihood alone, however, can be deceptive, *i.e.*, it may appear close to optimal even though the model structure represents poor generalization. This is because the critical, longer samples that would be indicative of generalization have small probability and contribute little to the average log likelihood. This was the primary reason for devising the parsing experiments as an additional evaluation criterion.

**Results** The merging procedure was able to find the target model structure for both types of training sets. The left-most data points in the plots can therefore be taken as benchmarks in evaluating the performance of the Baum-Welch method on this data.

The quality of the Baum-Welch induced model structures seems to vary wildly with the choice of initial conditions. For the minimal sample, 2 out of 10 runs resulted in perfect model structures when working with 6 states; 3 out of 10 when using 10 states (four more than necessary). When given a random training sample instead, the success rate improved to 3/10 and 7/10, respectively.

The overgeneralizations observed in Baum-Welch derived models correspond mostly to a missing correlation between initial and final symbols. These models typically generate some subset of $(a \cup b)c^*(a \cup b)$ which leads to about 50% of the samples generated to be rejected by the target model (cf. bottom plots).

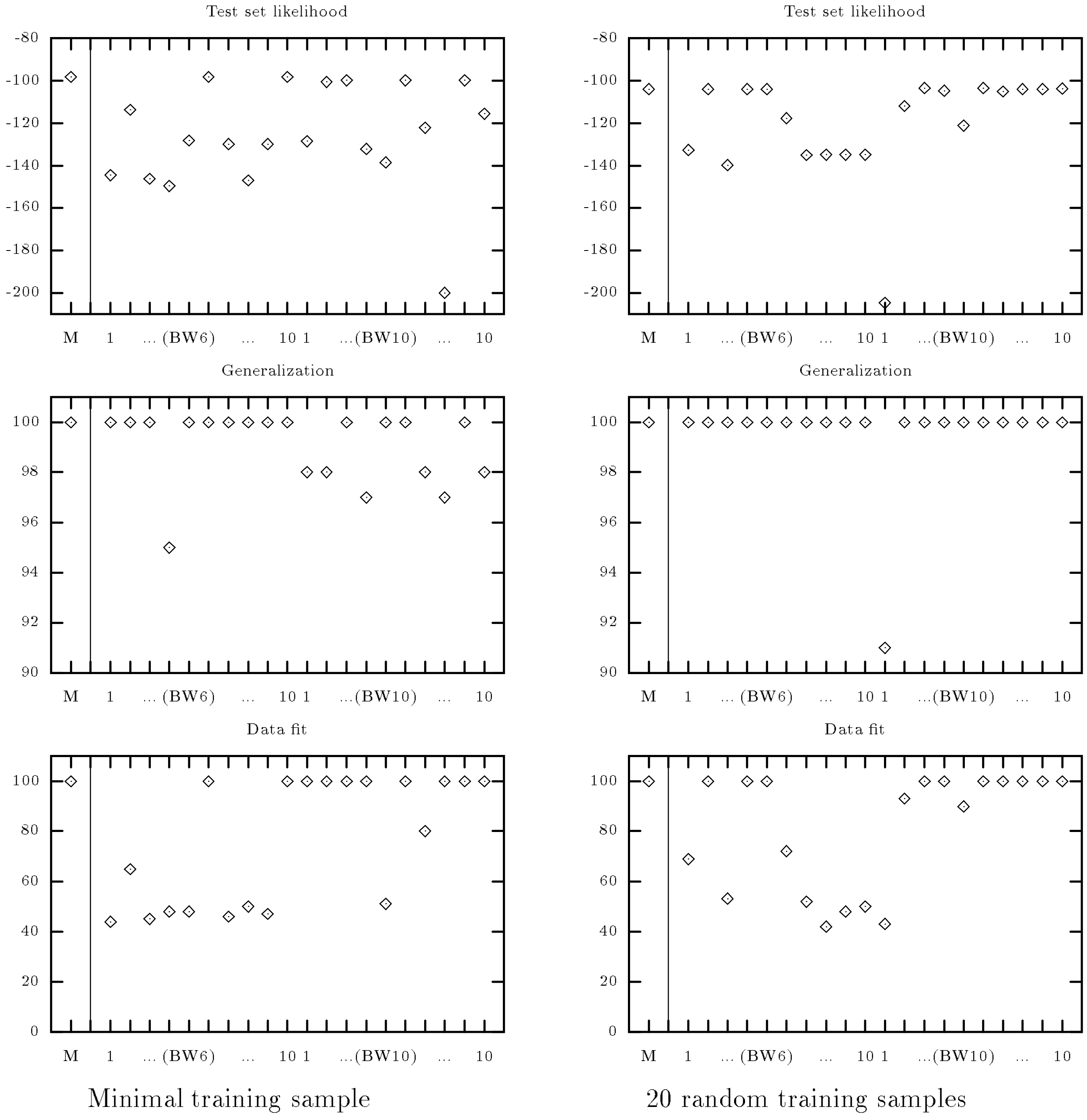

Figure 6: Case study I: Results of induction runs.

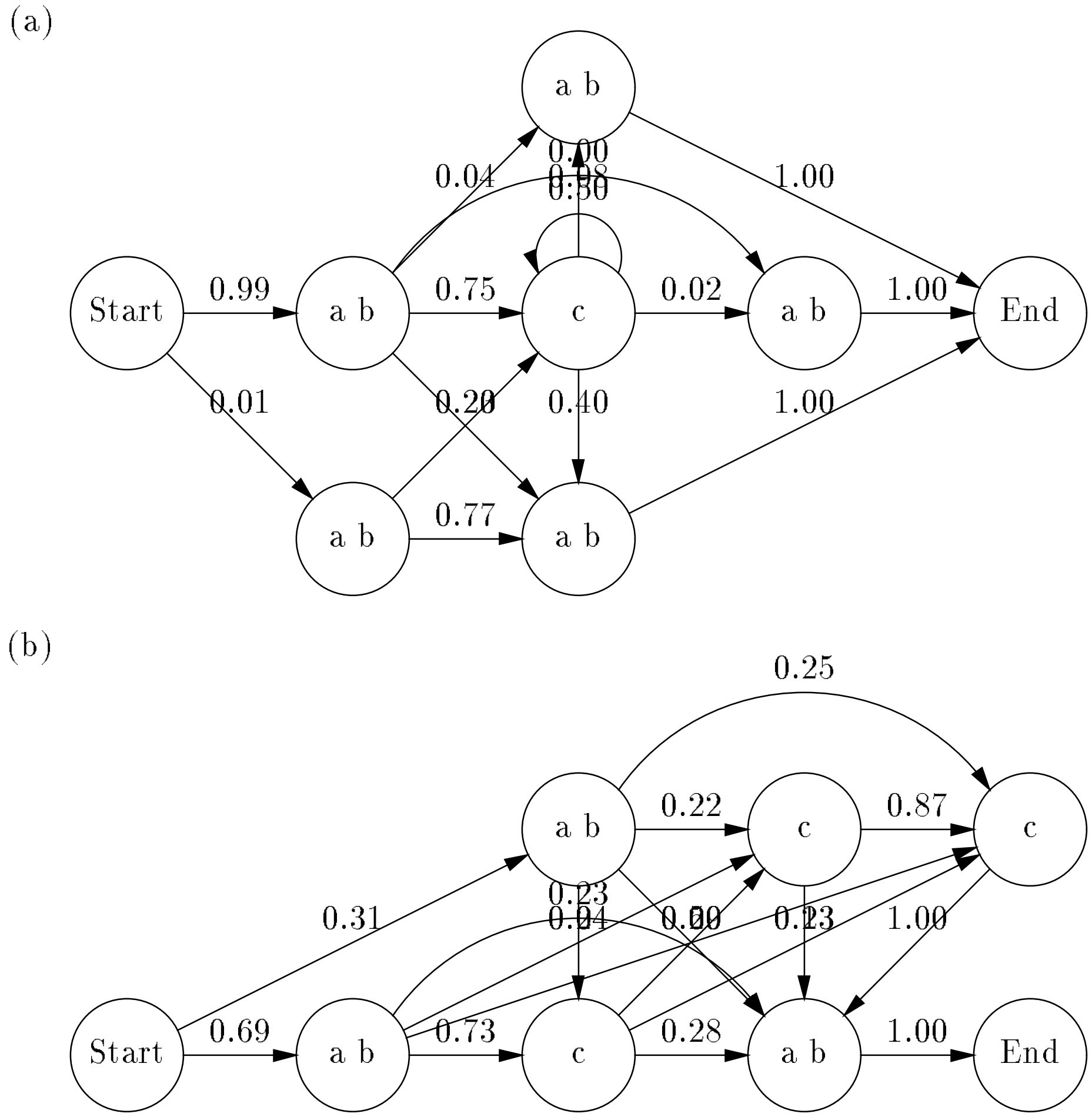

Figure 7: Case study I: BW-derived HMM structures that fail on generalization.

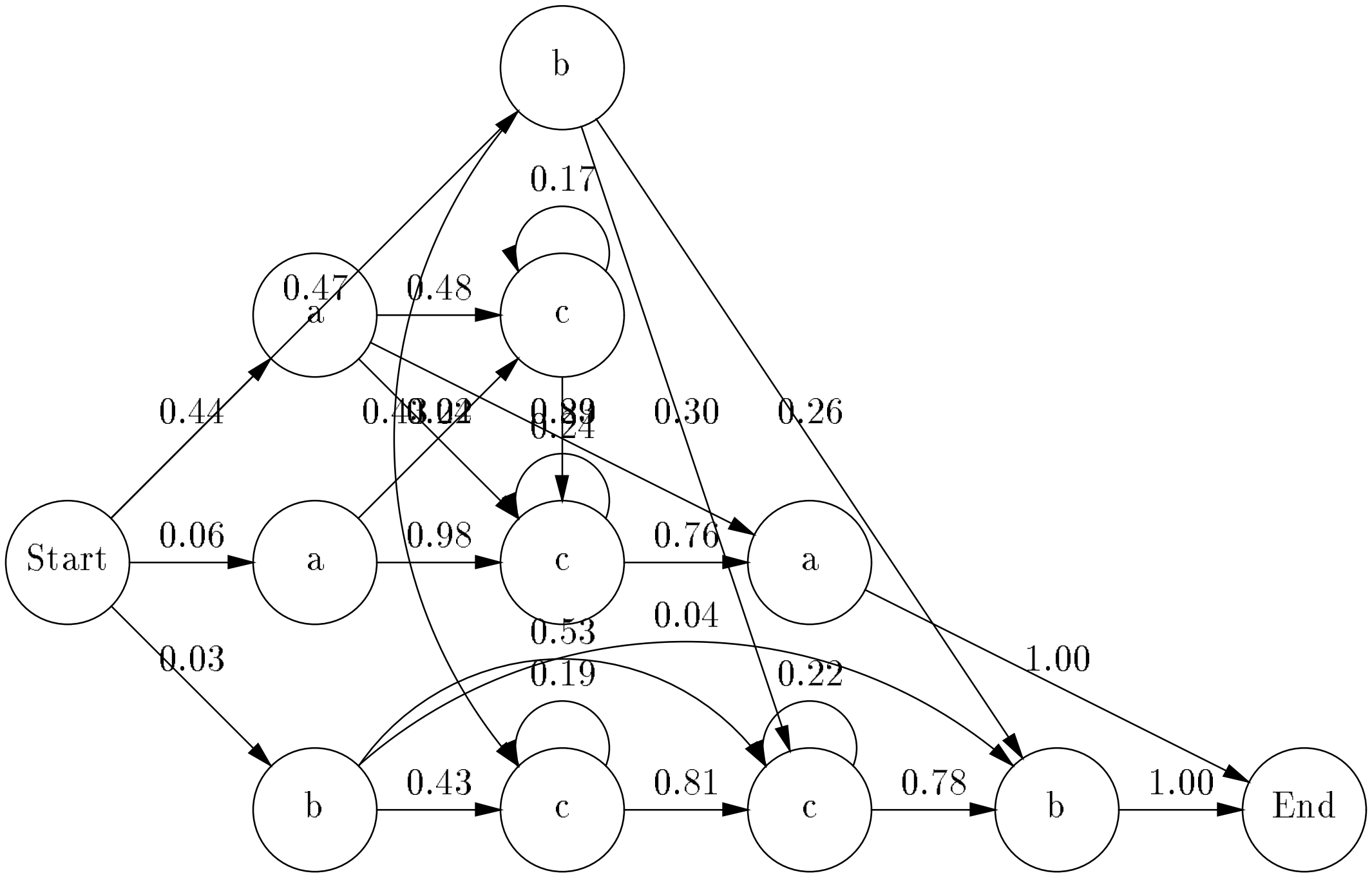

Figure 8: Case study I: Redundant BW-derived HMM structure for $ac^*a \cup bc^*b$.

**Baum-Welch studies** It is instructive to inspect some of the HMM topologies found by the Baum-Welch estimator. Figure 7 shows models of 6 states trained on minimal samples, one exhibiting overgeneralization, and one demonstrating both overfitting and overgeneralization.

The HMM in (a) generates $(a \cup b)c^*(a \cup b)$ and has redundantly allocated states to generate $a \cup b$. The HMM in (b) generates $(a \cup b)c^k(a \cup b)$, for $k = 0, 1, 2, 3$. Here, precious states have been wasted modeling the repetition of $c$'s, instead of generalizing to a loop over a single state and using those states to model the distinction between $a$ and $b$.

If estimation using the minimal number of states (6 in this case) is successful, the discretized structure invariably is that of the target model (Figure 5), as expected, although the probabilities will depend on the training sample used. Successful induction using 10 states, on the other hand, leads to models that, by definition, contain redundant states. However, the redundancy is not necessarily a simple duplication of states found in the target model structure. Instead, rather convoluted structures are found, such as the one in Figure 8 (induced from the random 20 samples).

**Merging studies** We also investigated how the merging algorithm behaves for non-optimal values of the global prior weight $\lambda$. As explained earlier, this value is implicit in the number of 'effective' samples, the parameter that was maintained constant in all experiments, and which seems to be robust over roughly an order of magnitude.

We therefore took the resulting $\lambda$ value and adjusted it both upward and downward by an order of magnitude to produce undergeneralized (overfitted) and overgeneralized models, respectively. The series of models found (using the minimal sample) is shown in Figure 9.

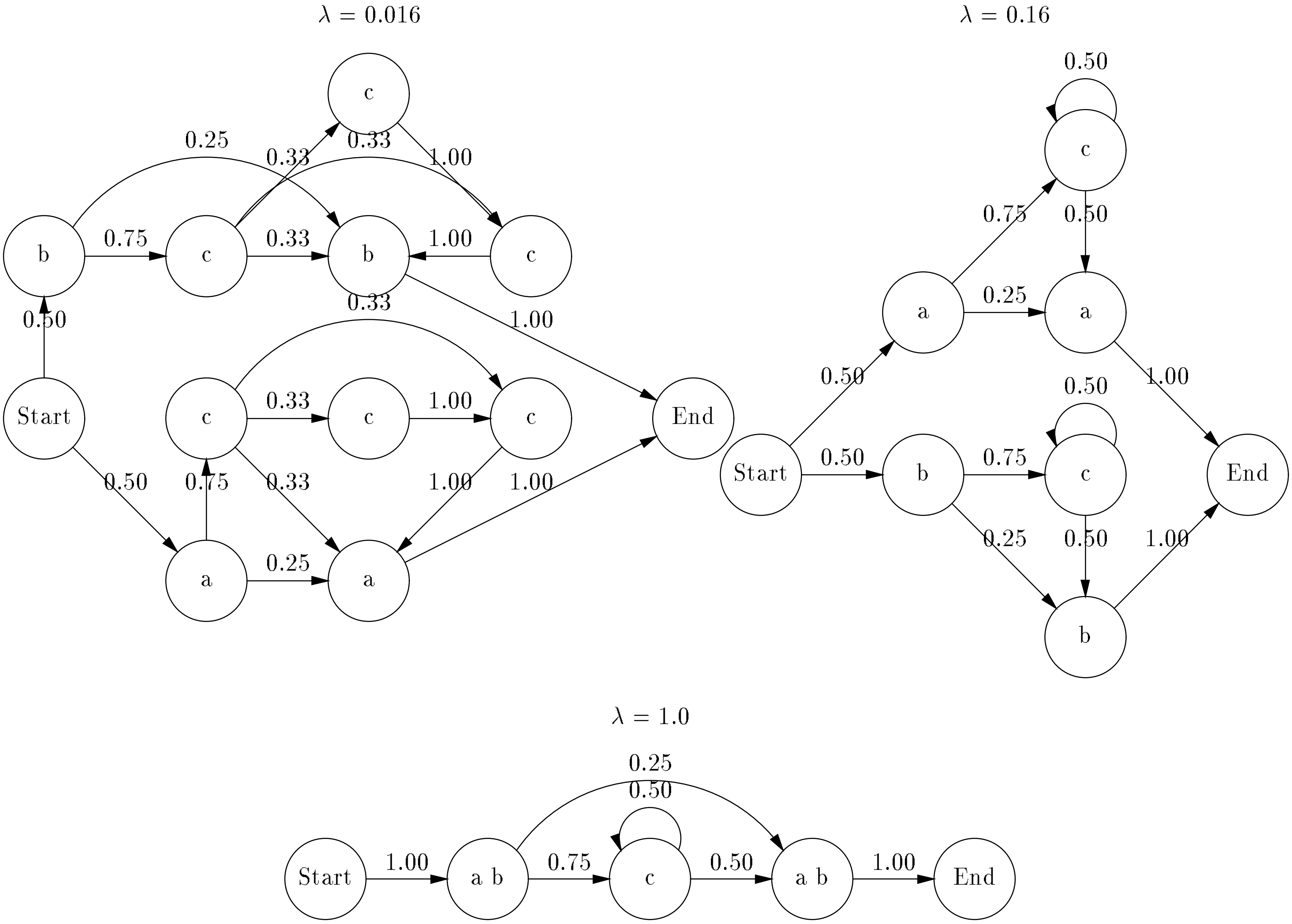

Figure 9: Case study I: Generalization depending on global prior weight.

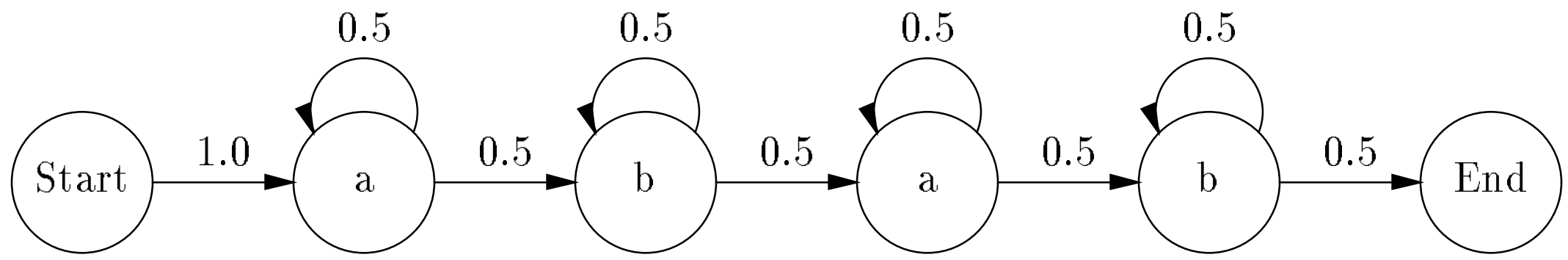

Figure 10: Case study II: HMM generating the test language $a^+b^+a^+b^+$.

For $\lambda = 0.016$ no structural generalization takes place; the sample set is simply represented in a concise manner. For a wide range around $\lambda = 0.16$, the target HMM is derived, up to different probability parameters. A further increase to $\lambda = 1.0$ produces a model whose structure no longer distinguishes between $a$ and $b$. One could argue that this overgeneralization is a 'natural' one given the data.

### 6.1.4 Case study II

The second test language is $a^+b^+a^+b^+$, generated by the HMM depicted in Figure 10. The minimal training sample contains the following nine strings

*abab*
*aabab*
*abbab*
*abaab*
*ababb*
*aaabab*
*abbbab*
*abaaab*
*ababbb*

The other training sample once again consisted of 20 randomly drawn strings.

Figure 11 presents the results in graphical form, using the same measures and arrangement as in the previous case study. (However, note that the ranges on some of the $y$-axes differ.)

Similar to the previous experiment, the merging procedure was successful in finding the target model, whereas the Baum-Welch estimator produced inconsistent results that were highly dependent on the initial parameter settings. Furthermore, the Baum-Welch success rates seemed to reverse when switching from the minimal to the random sample (from 6/10 and 0/10 to 1/10 and 6/10, respectively). This is disturbing since it reveals a sensitivity not only to the number of states in the model, but also to the precise statistics of the sample data.

The overgeneralizations are typically of the form $(a^+b^+)^k$, where either $k = 1$ or $k > 2$.

**Baum-Welch studies** As in the previous case study, we looked at various model structures found by Baum-Welch estimation. All examples in this section are from training on the 20 random samples.

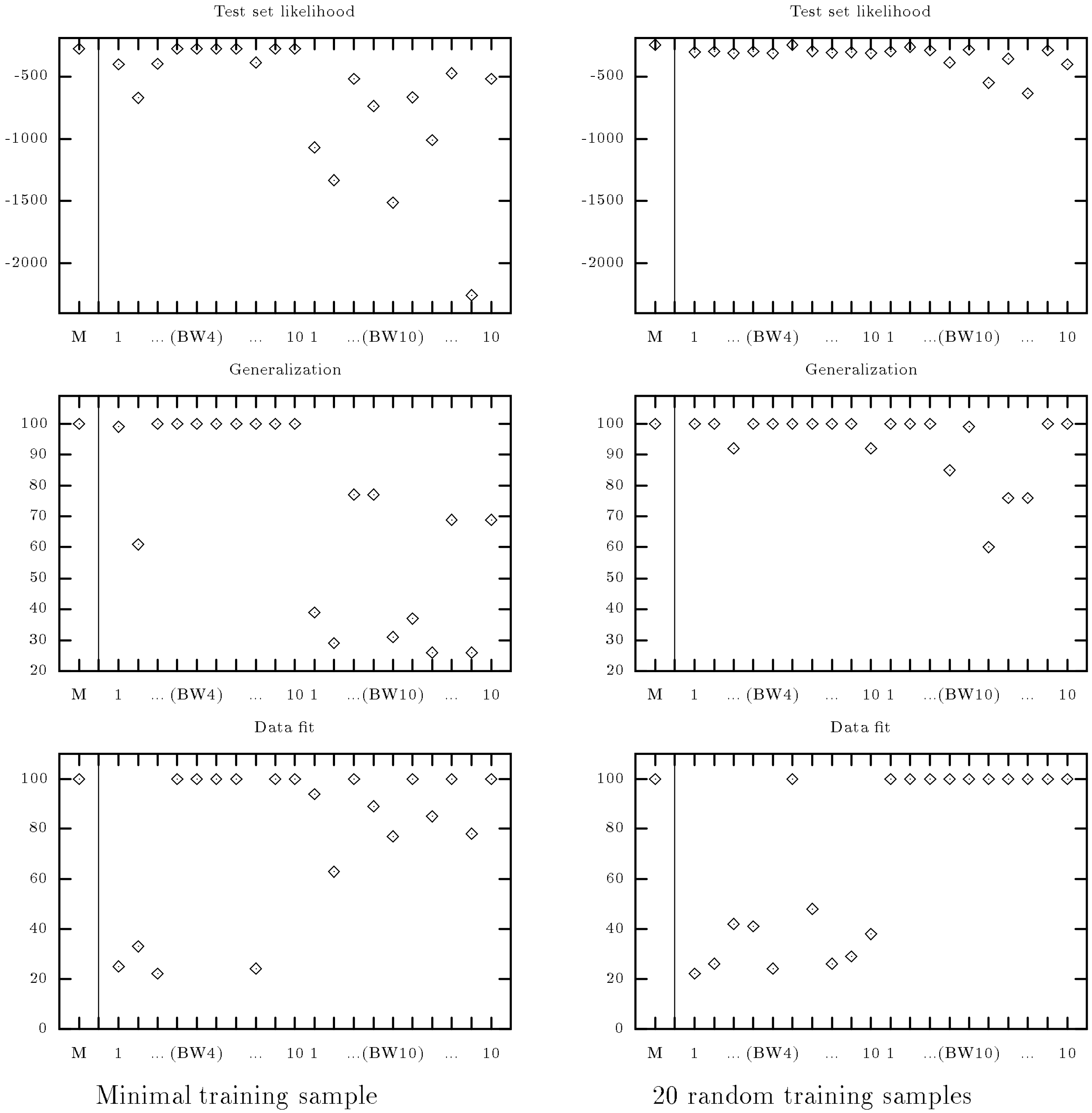

Figure 11: Case study II: Results of induction runs.

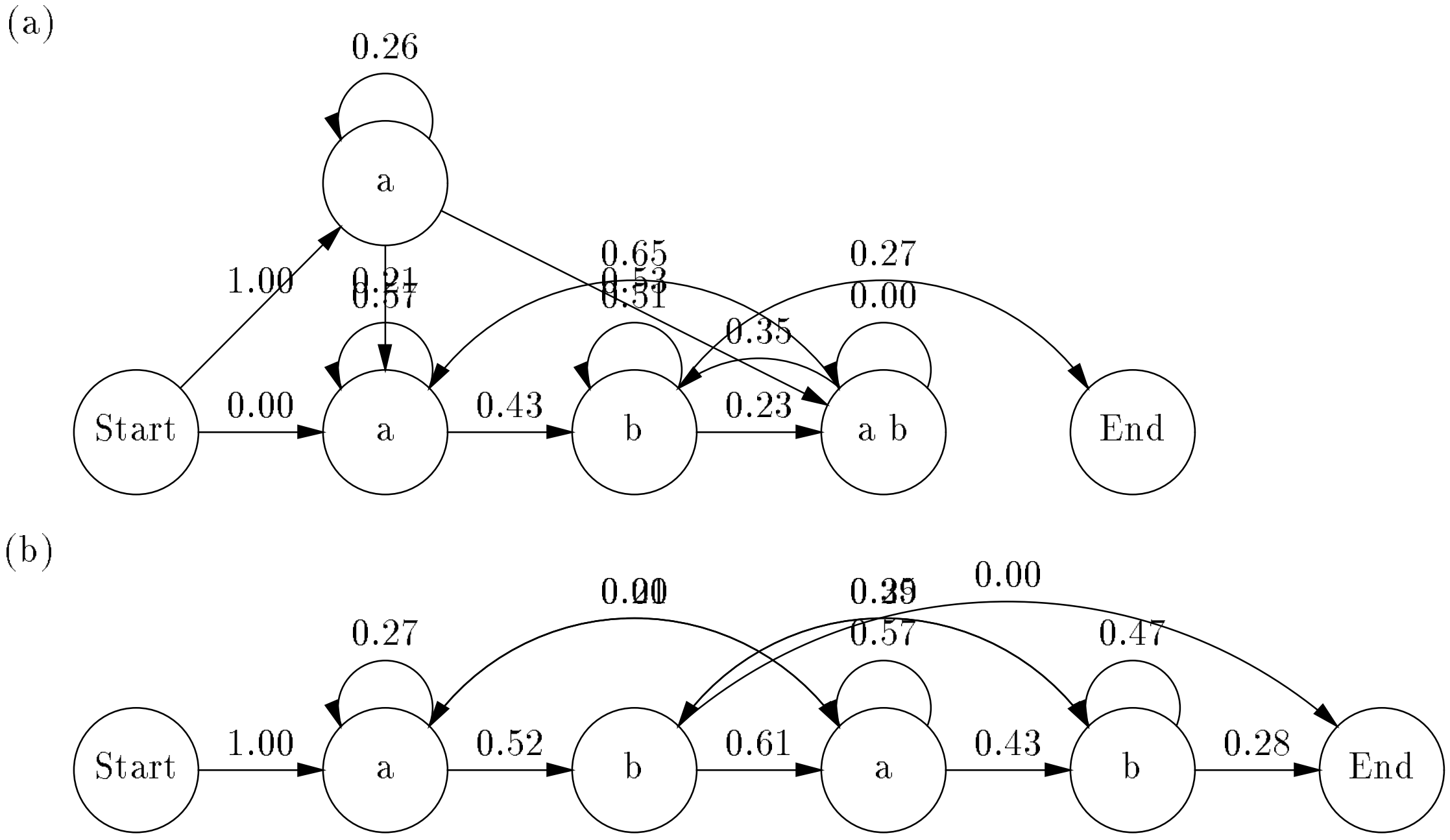

Figure 12: Case study II: BW-derived HMM structures that fail on generalization.

Figure 12(a) shows a structure that is overly general: it generates $(a \cup a^+b^+)(a \cup b)^+b^+$. In (b), we have an HMM that partly overgeneralizes, but at the same time exhibits a rather peculiar case of overfitting: it excludes strings of the form $a^+b^ka^+b^+$ where $k$ is even. No such cases happened to be present in the training set.

The accurate model structures of 10 states found by the Baum-Welch method again tended to be rather convoluted. Figure 13 shows as case in point.

**Merging studies** We also repeated the experiment examining the levels of generalization by the merging algorithm, as the value of the global prior weight was increased over three orders of magnitude.

Figure 14 shows the progression of models for $\lambda = 0.018, 0.18$, and 1.0. The pattern is similar to that in in the first case study (Figure 14). The resulting models range from a simple merged representation of the samples to a plausible overgeneralization from the training data ($(a^+b^+)^+$). The target model is obtained for $\lambda$ values between these two extremes.

### 6.1.5 Discussion

It is tempting to try to find a pattern in the performance of the Baum-Welch estimator in terms of parameters such as model size and sample size and type (although this would go beyond the intended scope of this study). Regarding model size, one would expect the smaller minimal models to produce better coverage of the target language, and a tendency to overgeneralize, since too few states are available to produce a close fit to the data. This is indeed observable in the plots in the bottom rows of Figures 6 and 11: the runs in the left half typically produce a higher number of rejected (overgeneralized) samples.

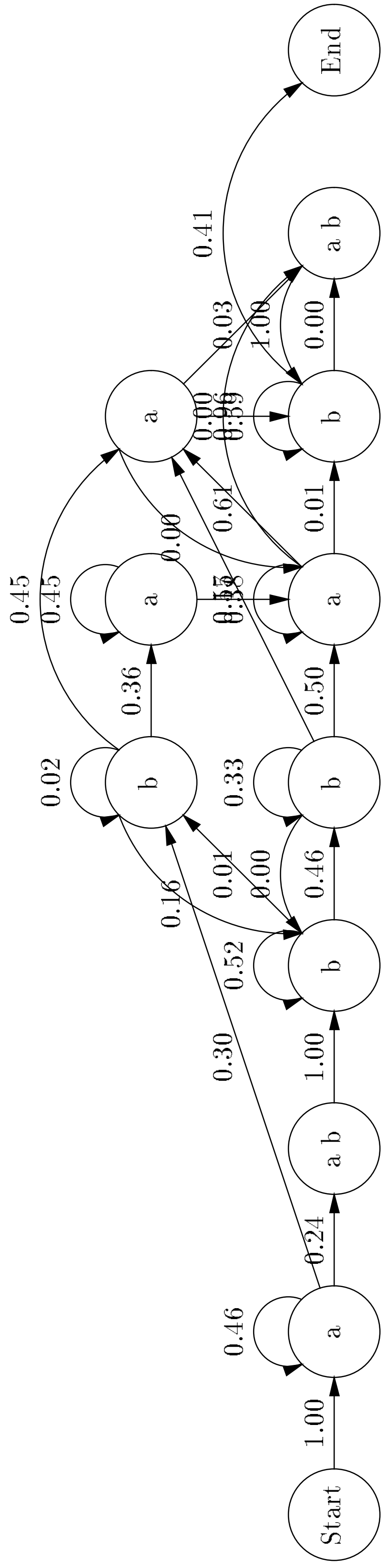

Figure 13: Case study II: Redundant BW-derived HMM structure for $a^+b^+a^+b^+$.

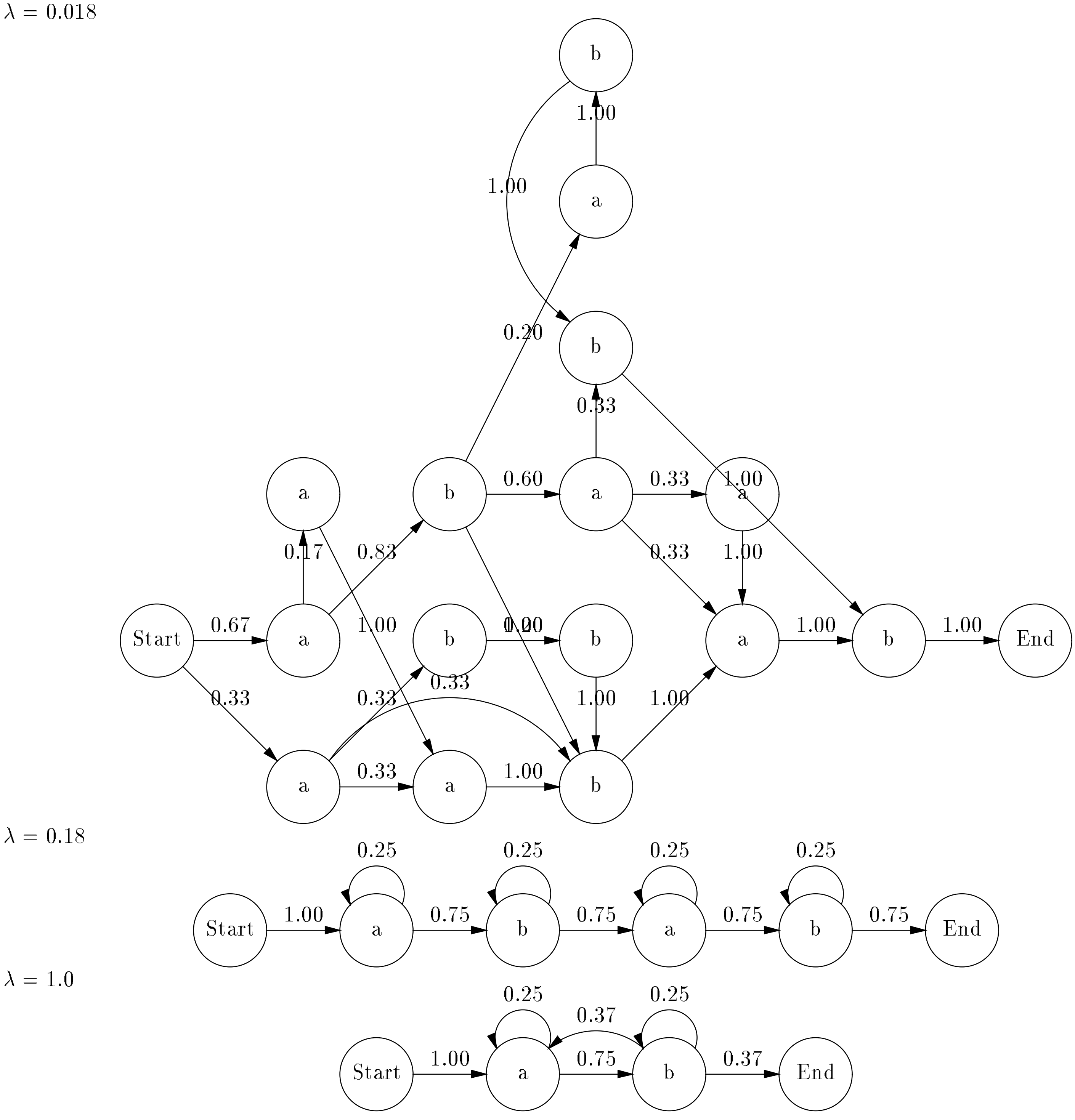

Figure 14: Case study II: Generalization depending on global prior weight.

Conversely, one expects a greater tendency toward overfitting in the training runs using more than the minimal number of states. The plots in the middle rows of Figures 6 and 11 confirm this expectation: the right halfs show a greater number of rejected strings from the target language, indicating insufficient generalization.

It is conceivable that for each language there exists a model size that would lead to a good compromise between generalization and data fit so as to produce reliable structure estimation. The problem is that there seems to be no good way to predict that optimal size.[13]

Successful use of the model merging approach also relies on suitable parameter choices, mainly of the global prior weight (or the number of 'effective samples'). The prime advantage of merging in this regard is that the parameters seem to be more robust to both sample size and distribution, and the mechanics of the algorithm make it straightforward to experiment with them. Furthermore, it appears that overgeneralization by excessive merging tends to produce 'plausible' models (with the obvious caveat that this conclusion is both tentative given the limited scope of the investigation, and a matter of human judgement).

## 6.2 Phonetic word models from labeled speech

### 6.2.1 The TIMIT database

In the second evaluation stage, we were looking for a sizeable collection of real-world data suitable for HMM modeling. The TIMIT (Texas Instruments-MIT) database is a collection of hand-labeled speech samples compiled for the purpose of training speaker-independent phonetic recognition systems (Garofolo 1988). It contains acoustic data segmented by words and aligned with discrete labels from an alphabet of 62 phones. For our purposes, we ignored the continuous, acoustic data and viewed the database simply as a collection of string samples over a discrete alphabet.

The goal is to construct a probabilistic model for each word in the database, representing its phonetic structure as accurately as possible, *i.e.*, maximizing the probabilities of the observed pronunciations. A fraction of the total available data, the *test set*, is set aside for evaluating the induced models, while the rest is used to induce or otherwise estimate the probabilistic model for each word. By comparing the performance of the models generated by various methods, along with other relevant properties such as model size, processing time, *etc.*, we can arrive at a reasonably objective comparison of the various methods. Of course, the ultimate test is to use pronunciation models in an actual system that handles acoustic data, a considerably more involved task. In Section 6.3 we will describe such a system and how the HMM merging process was brought to bear on it.

The full TIMIT dataset consists of 53355 phonetic samples for 6100 words.[14] To keep the task somewhat manageable, and to eliminate a large number of words with too few samples to allow meaningful structural model induction, we used a subset of this data consisting of words of intermediate frequency. Arbitrarily, we included words occurring between 20 and 100 times in the dataset. This left us with a working dataset of 206 words, comprising a total of 7861 samples. Of these, 75% for each word (5966 total) were made available to various training methods, while the remaining 25% (1895 total) were left for evaluation.

[13] In Section 6.2 we actually use a simple heuristic that scales the model sizes linearly with the length of the samples. This heuristic works rather well in that particular application, but it crucially relies on the models being loop-free, and hence wouldn't apply generally.

[14] This is the union of the 'training' and 'test' portions in the original TIMIT distribution.

### 6.2.2 Qualitative evaluation

In preliminary work, while evaluating the possibility of incorporating HMM merging in an ongoing speech understanding project, Gary Tajchman, a researcher at ICSI with extensive experience in acoustic-phonetic modeling, experimented with the algorithm using the TIMIT database. He inspected a large number of the resulting models for 'phonetic plausibility', and found that they generally appeared sound, in that the generated structures were close to conventional linguistic wisdom.

To get an idea for the kinds of models that HMM merging produces from this data it is useful to examine an example. Figure 16 shows an HMM constructed from 37 samples of the word *often*. For comparison, Figure 15 shows the initial HMM constructed from the samples, before merging (but with identical paths collapsed).

Figure 17 plots the number of states obtained during on-line (incremental) merging as a function of the number of incorporated samples. The numbers of states before and after merging at each stage are plotted as adjacent datapoint, giving rise to the spikes in the figure. Merging starts with five incorporated samples (17 states). Initially merging occurs with almost every additional samples, but later on most samples are already parsed by the HMM and warrant no further merging.

The most striking feature of the induced HMM is that both the first and the second syllable of *often* contain a number of alternative pronunciations anchored around the central [f] consonant, which is common to all variants. This structural property is well mirrored in the two branching sections of the HMM. A number of the pronunciation for the second syllable share the optional [tcl t] sequence.

Notice that each state (except initial and final) has exactly one output symbol. This constraint was imposed due to the particular task we had in mind for the resulting HMMs. The speech recognition system for which these word models are intended implicitly assumes that each HMM state represents a unique phone. Such a restriction can be easily enforced in the algorithm by filtering the merge candidates for pairs of states with identical emissions.

The single-output constraint does not limit the representational power of the HMMs, since a multi-output state can always be split into several single-output states. It does, however, affect the structural prior for the HMM. In a single-output HMM, each emission carries a prior probability of $\frac{1}{|\Sigma|}$, rather than one of the various structural priors over multinomials discussed in Section 3.4.3. Incidentally, the constraint can also speed up the algorithm significantly, because many candidates are efficiently eliminated that would otherwise have to be evaluated and rejected. This advantage by far outweighs the larger size of the derived models.

A second constraint can be enforced in this particular domain (and possibly others). Since the resulting HMMs are meant to be phonetic word models, it does not make sense to allow loops in these models. In very rare circumstances the merging algorithm might be tempted to introduce such loops, *e.g.*, because of some peculiar repetitive pattern in

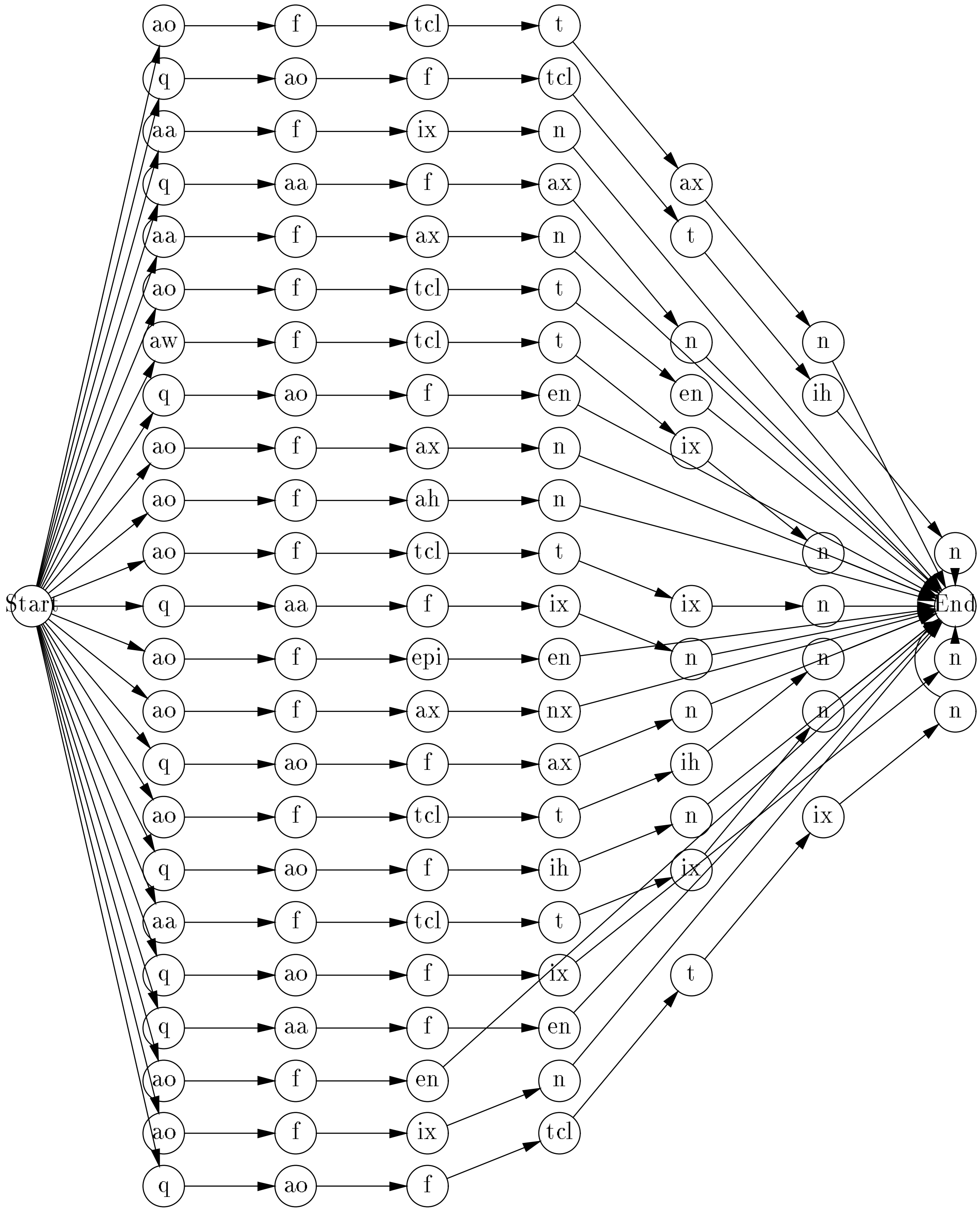

Figure 15: Initial HMM constructed from 37 samples of the word *often*.

Probabilities are omitted in this graph. Due to repetitions in the data the HMM has only 23 distinct paths.

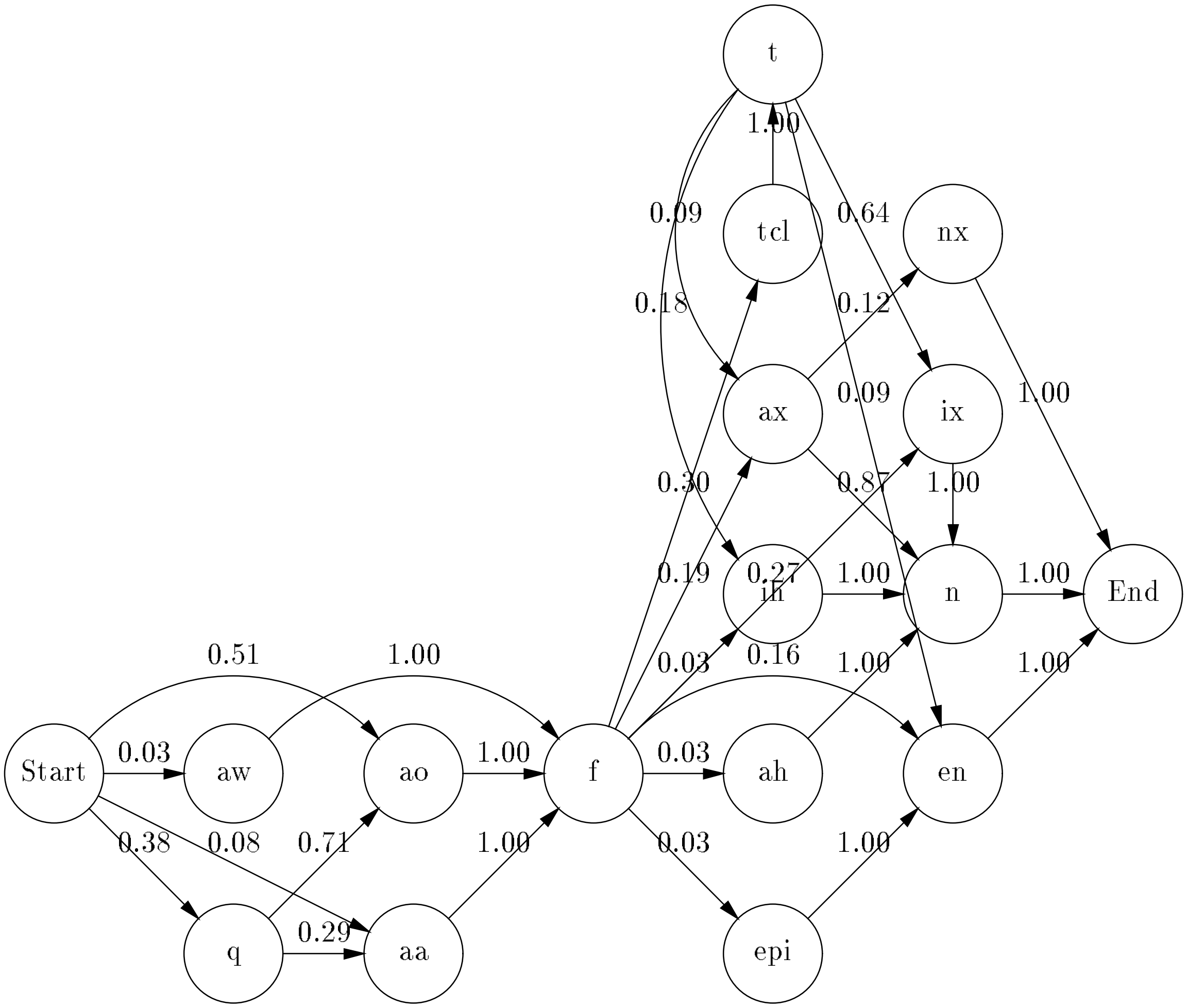

Figure 16: Merged HMM constructed from 37 samples of the word *often*.

Merging was constrained to keep the emission on each state unique.

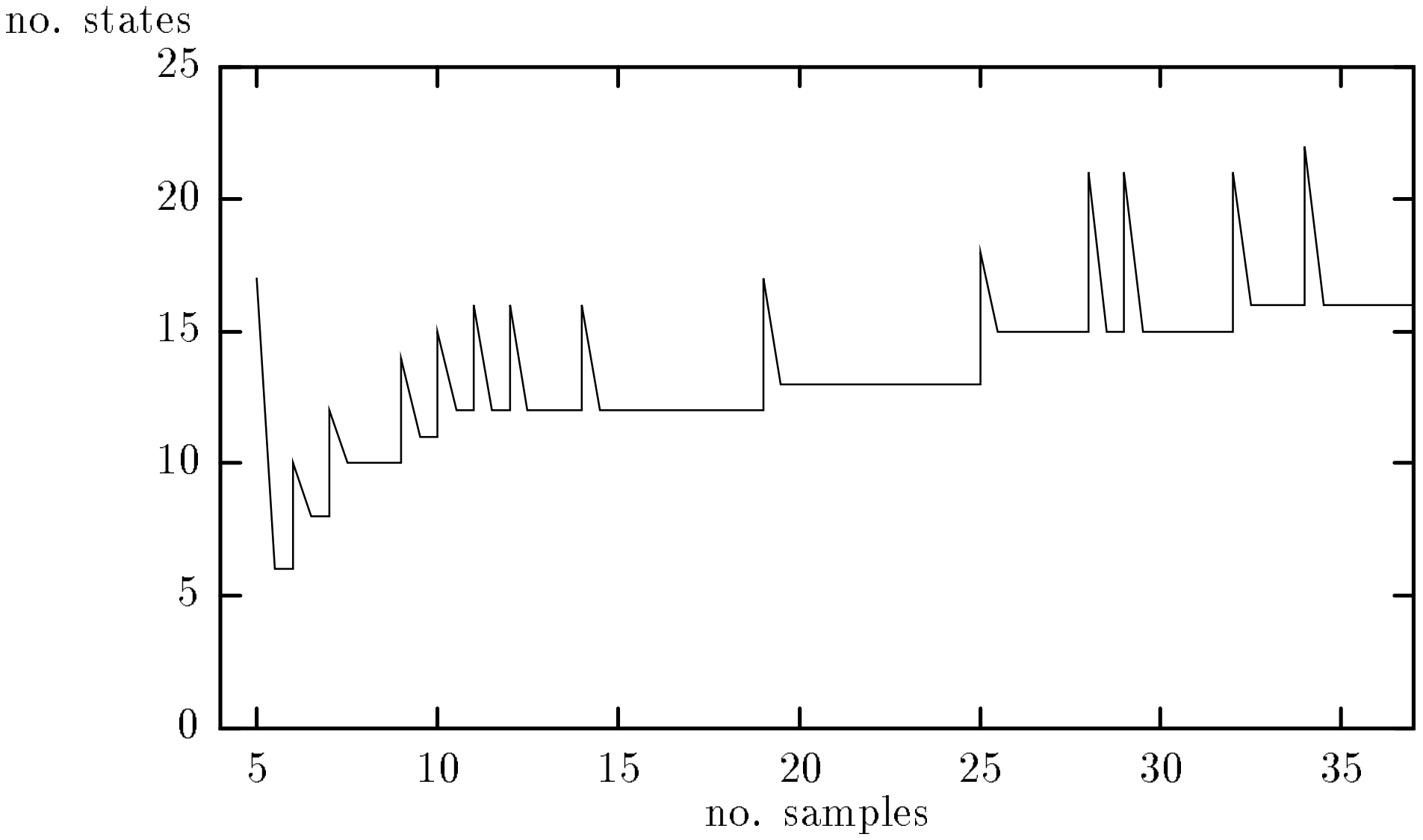

Figure 17: Number of HMM states as a function of the number of samples incorporated during incremental merging.

Each spike represents the states added to model an unparseable sample, which are then (partly) merged into the existing HMM structure.

a sample ('banana'). Given our prior knowledge in this regard we can simply rule out candidate merges that would introduce loops.[15]

HMM merging can thus be used to derive models for allophonic variation from data, without explicitly representing a mapping from individual phonemes to their realizations. This is in contrast to other approaches where one first induces rules for pronunciations of individual phonemes based on their contexts (*e.g.*, using decision tree induction), which can then be concatenated into networks representing word pronunciations (Chen 1990; Riley 1991). A detailed comparison of the two approaches is desirable, but so far hasn't been carried out. We simply remark that both approaches could be combined by generating allophone sequences from induced phoneme-based models and adding them to directly observed pronunciations for the purpose of smoothing.

### 6.2.3 Quantitative evaluation

Several HMM construction methods were tested and compared using the TIMIT data.

- The ML model: the HMM is the union of all unique samples, with probabilities corresponding to the observed relative frequencies. This is essentially the result of building the initial model in the HMM merging procedure, before any merging takes place.

- Baum-Welch estimation: an HMM of fixed size and structure is submitted to Baum-Welch (EM) estimation of the probability parameters. Of course, finding the 'right' size and structure is exactly the learning problem at hand. We wanted to evaluate the structure finding abilities of the Baum-Welch procedure, so we set the number of states to a fixed multiple of the maximum sample length for a given word, and randomly initialized all possible transitions and emissions to non-zero probabilities. After the EM procedure converges, transitions and emissions with probability close to zero are pruned, leaving an HMM structure that can be evaluated. Several model sizes (as multiples of the sample length) were tried.

- Standard HMM merging with loop suppression (see above). We used the simple description length prior from Section 3.4.3, with $\log P = -n_t \log |Q|$ for transitions and $\log P = -n_e \log |\Sigma|$ for emissions, as well as a narrow Dirichlet prior for transition and emission probabilities, with $\alpha_0 = 1.0$ in both cases. Several global weighting factors $\lambda$ for the structure prior were evaluated.

- HMM merging with the single-output constraint, as explained above. The same prior as for the multiple-output HMMs was used, except that an emission parameter prior does not exist in this case, and the structural prior contribution for an emission is $\log P = -\log |\Sigma|$.

[15] It is customary in HMM modeling for speech recognition to introduce *self-loops* on states to model varying durations. This does not contradict what was said above; these self-loops are introduced at a lower representational level, along with other devices such as state replication. They are systematically added to the merged HMM before using the HMM for alignment with continuous speech data.

A simple-minded way of comparing these various methods would be to apply them to the training portion of the data, and then compare generalization on the test data. As a measure of generalization it is customary to use the negative log probability, or empirical cross-entropy, they assign to the test samples. The method that achieves the lowest cross-entropy would 'win' the comparison.

A problem that immediately poses itself is that there is a significant chance that some of the test samples have zero probability according to the induced HMMs. One might be tempted to evaluate based on the number of test samples covered by the model, but such a comparison alone would be meaningless since a model that assigns (very low) probability to all possible strings could trivially 'win' in this comparison.

The general approach that is usually taken in this situation is to have some recipe that prevents vanishing probabilities on new, unseen samples. There are a great many such approaches in common use, such as parameter smoothing and back-off schemes, but many of these are not suitable for the comparison task at hand.

At the very least, the method chosen should be

- well-defined, *i.e.*, correspond to some probabilistic model that represents a proper distribution over all strings;

- unbiased with respect to the methods being compared, to the extent possible.

Back-off models (where a second model is consulted if, and only if, the first one returns probability zero) do not yield consistent probabilities unless they are combined with 'discounting' of probabilities to ensure that the total probability mass sums up to unity (Katz 1987). The discounting scheme, as well as various smoothing approaches (*e.g.*, adding a fixed number of virtual 'Dirichlet' samples into parameter estimates) tend to be specific to the model used, and are therefore inherently problematic when comparing different model-building methods.

To overcome these problems, we chose to use *mixture models*. The target models to be evaluated are combined with a simple back-off model that guarantees non-zero probabilities, *e.g.*, a bigram grammar with smoothed parameters. This back-off grammar is identical in structure for all target models. Unlike discrete back-off schemes, the target and the back-up are always consulted both for the probability they assign to a given sample, which are then weighted and averaged according to a mixture proportion. The resulting overall model is probabilistically sound: it corresponds to a stochastic process that first flips a coin to determine which of the two components should generate the sample, and then dispatches sample generation to the winner of that coin flip.

When comparing two model induction methods, we first let each induce a structure. Each is built into a mixture model, and both the component model parameters and the mixture proportions are estimated using a variant of the EM procedure for generic mixture distributions (Redner & Walker 1984). To get meaningful estimates for the mixture proportions, the HMM structure should be induced based on a subset of the training data, and the full training data is then used to estimate the parameters, including the mixture weights. This holding-out of training data makes the mixture model approach similar to the deleted interpolation method (Jelinek & Mercer 1980). The main difference is that the component parameters are estimated jointly with the mixture proportions.[16] In our experiments we always used half of the training data in the structure induction phase, adding the other half during the EM estimation phase.

### 6.2.4 Results and discussion

HMM merging was evaluated in two variants, with and without the single-output constraint. In each version, three settings of the structure prior weight $\lambda$ were tried: 0.25, 0.5 and 1.0. Similarly, for Baum-Welch training the preset number of states in the fully parameterized HMM was set to 1.0, 1.5 and 1.75 times the longest sample length. For comparison purposes, we also included the performance of the unmerged maximum-likelihood HMM, and a biphone grammar of the kind used in the mixture models used to evaluate the other model types. Table 1 summarizes the results of these experiments.

The table also includes useful summary statistics of the model sizes obtained, and the time it took to compute the models. The latter figures are obviously only a very rough measure of computational demands, and their comparison suffers from the fact that the implementation of each of the methods may certainly be optimized in idiosyncratic ways. Nevertheless these figures should give an approximate idea of what to expect in a realistic application of the induction methods involved.[17]

One important general conclusion from these experiments is that both the merged models and those obtained by Baum-Welch training do significantly better than the two 'dumb' approaches, the bigram grammar and the ML HMM (which is essentially a list of observed samples). We can therefore conclude that it pays to try to generalize from the data, either using our Bayesian approach or Baum-Welch on an HMM of suitable size.

Overall the difference in scores even between the simplest approach (bigram) and the best scoring one (merging, $\lambda = 1.0$) are quite small, with phone perplexities ranging from 1.985 to 1.849. This is not surprising given the specialized nature and small size of the sample corpus. Unfortunately, this also leaves very little room for significant differences in comparing alternate methods. However, the advantage of the best model merging result (unconstrained outputs with $\lambda = 1.0$) is still significant compared to the best Baum-Welch (size factor 1.5) result ($p < 0.041$). Such small differences in log probabilities would probably be irrelevant when the resulting HMMs are embedded in a speech recognition system.

Perhaps the biggest advantage of the merging approach in this application is the compactness of the resulting models. The merged models are considerably smaller than the comparable Baum-Welch HMMs. This is important for any of the standard algorithms operating on HMMs, which typically scale linearly with the number of transitions (or quadratically with the number of states). Besides this advantage in production use, the training times for Baum-Welch grow quadratically with the number of states for the structure induction phase since it requires fully parameterized HMMs. This scaling is clearly visible in the run times we observed.

[16] This difference can be traced to the different goals: in deleted interpolation the main goal is to gauge the reliability of parameter estimates, whereas here we want to assess the different structures.

[17] All evaluations were carried out using a very flxible, but not necessarily fast, system for object-oriented experimentation with grammars and grammar induction, written in Common Lisp and CLOS. Interested researchers should contact the first author for access to this code.

| | ML | M ($\lambda = 0.25$) | M ($\lambda = 0.5$) | M ($\lambda = 1.0$) |
|---|---|---|---|---|
| $\log P$ | $-2.600 \cdot 10^3$ | $-2.418 \cdot 10^3$ | $-2.355 \cdot 10^3$ | $-2.343 \cdot 10^3$ |
| Perplexity | 1.979 | 1.886 | 1.855 | 1.849 |
| Significance | $p < 0.000001$ | $p < 0.0036$ | $p < 0.45$ | – |
| states | 4084 | 1333 | 1232 | 1204 |
| transitions | 4857 | 1725 | 1579 | 1542 |
| emissions | 3878 | 1425 | 1385 | 1384 |
| training time | 28:19 | 32:03 | 28:58 | 29:49 |

| | ML | M1 ($\lambda = 0.25$) | M1 ($\lambda = 0.5$) | M1 ($\lambda = 1.0$) |
|---|---|---|---|---|
| $\log P$ | $-2.600 \cdot 10^3$ | $-2.450 \cdot 10^3$ | $-2.403 \cdot 10^3$ | $-2.394 \cdot 10^3$ |
| Perplexity | 1.979 | 1.902 | 1.879 | 1.874 |
| Significance | $p < 0.000001$ | $p < 0.0004$ | $p < 0.013$ | $p < 0.016$ |
| states | 4084 | 1653 | 1601 | 1592 |
| transitions | 4857 | 2368 | 2329 | 2333 |
| emissions | 3878 | 1447 | 1395 | 1386 |
| training time | 28:19 | 30:14 | 26:03 | 25:53 |

| | BG | BW ($N = 1.0L$) | BW ($N = 1.5L$) | BW ($N = 1.75L$) |
|---|---|---|---|---|
| $\log P$ | $-2.613 \cdot 10^3$ | $-2.470 \cdot 10^3$ | $-2.385 \cdot 10^3$ | $-2.392 \cdot 10^3$ |
| Perplexity | 1.985 | 1.912 | 1.870 | 1.873 |
| Significance | $p < 0.000001$ | $p < 0.000003$ | $p < 0.041$ | $p < 0.017$ |
| states | n/a | 1120 | 1578 | 1798 |
| transitions | n/a | 1532 | 2585 | 3272 |
| emissions | n/a | 1488 | 1960 | 2209 |
| training time | 3:47 | 55:36 | 99:55 | 123:59 |

Table 1: Results of TIMIT trials with several model building methods. The training methods are identified by the following keys: BG bigram grammar, ML maximum likelihood HMM, BW Baum-Welch trained HMM, M merged HMM, M1 single-output merged HMM. $\log P$ is the total log probability on the 1895 test samples. Perplexity is the average number of phones that can follow in any given context within a word (computed as the exponential of the per-phone cross-entropy). Significance refers to the $p$ level in a $t$-test pairing the log probabilities of the test samples with those of the best score (merging, $\lambda = 1.0$).

The number of states, transitions and emissions is listed for the resulting HMMs where applicable. The training times listed represent the total time (in minutes and seconds) it took to induce the HMM structure and subsequently EM-train the mixture models, on a SPARCstation 10/41.

Although we haven't done a word-by-word comparison of the HMM structures derived by merging and Baum-Welch, the summary of model sizes seem to confirm our earlier finding (Section 6.1) that Baum-Welch training needs a certain redundancy in 'model real estate' to be effective in finding good-fitting models. Smaller size factors give poor fits, whereas sufficiently large HMMs will tend to overfit the training data.

The choice of the prior weights $\lambda$ for HMM merging controls the model size in an indirect way: larger values lead to more generalization and smaller HMMs. For best results this value can be set based on previous experience with representative data. This could effectively be done in a cross-validation like procedure, in which generalization is successively increased starting with small $\lambda$'s. Due to the nature of the merging algorithm, this can be done incrementally, *i.e.*, the outcome of merging with a small $\lambda$ can be submitted to more merging at a larger $\lambda$ value, until further increases reduce generalization on the cross-validation data.

## 6.3 Multiple pronunciation word models for speech recognition

As part of his dissertation research, Wooters (1993) has used HMM merging extensively in the context of the Berkeley Restaurant Project (BeRP). BeRP is medium vocabulary, speaker-independent spontaneous continuous speech understanding system that functions as a consultant for finding restaurants in the city of Berkeley, California (Jurafsky *et al.* 1994).

In this application, the merging algorithm is run on strings of phone labels obtained by Viterbi-aligning previously existing word models to sample speech (using the TIMIT labels as the phone alphabet). As a result, new word models are obtained, which are then again used for Viterbi alignment, leading to improved labelings, *etc.* This procedure is iterated until no further improvement in the recognition performance (or the labelings themselves) is observed. The word models are bootstrapped with a list of pronunciations from a variety of databases. The goal of the iteration with repeated alignments and mergings is to tailor the word models to the task-specific data at hand, and to generalize from it where possible.

An added complication is that HMMs with discrete outputs are not by themselves applicable to acoustic speech data. Using an approach developed by Bourlard & Morgan (1993), the HMMs are combined with acoustic feature densities represented by a multi-layer perceptron (MLP). This neural network maps each frame of acoustic features into the phone alphabet. From the network outputs, the likelihoods of the HMM states relative to the acoustic emissions can be computed, as required for the Viterbi alignment or other standard HMM algorithms.

Since these emission probabilities are also subject to change due to changes of the word models, they too have to be reestimated on each iteration. The MLP is bootstrapped with weights obtained by training on the pre-labeled TIMIT acoustic data. Figure 18 depicts the combined iteration of MLP training, word model merging, and Viterbi alignment. It can be viewed as an instance of a generalized EM algorithm, in which emission probabilities (represented by the MLP) and HMM structure and transition probabilities are optimized separately. The separation is a result of the different technologies used.

For the BeRP system, HMM merging made it possible and practical to use multiple pronunciation word models, whereas before it was confined to single pronunciation models. (Note that in this setting, even a very restricted HMM can produce any acoustic emission with non-zero probability, due to the continuous nature of the domain, and because the emission distribution represented by the MLP is inherently non-vanishing.)

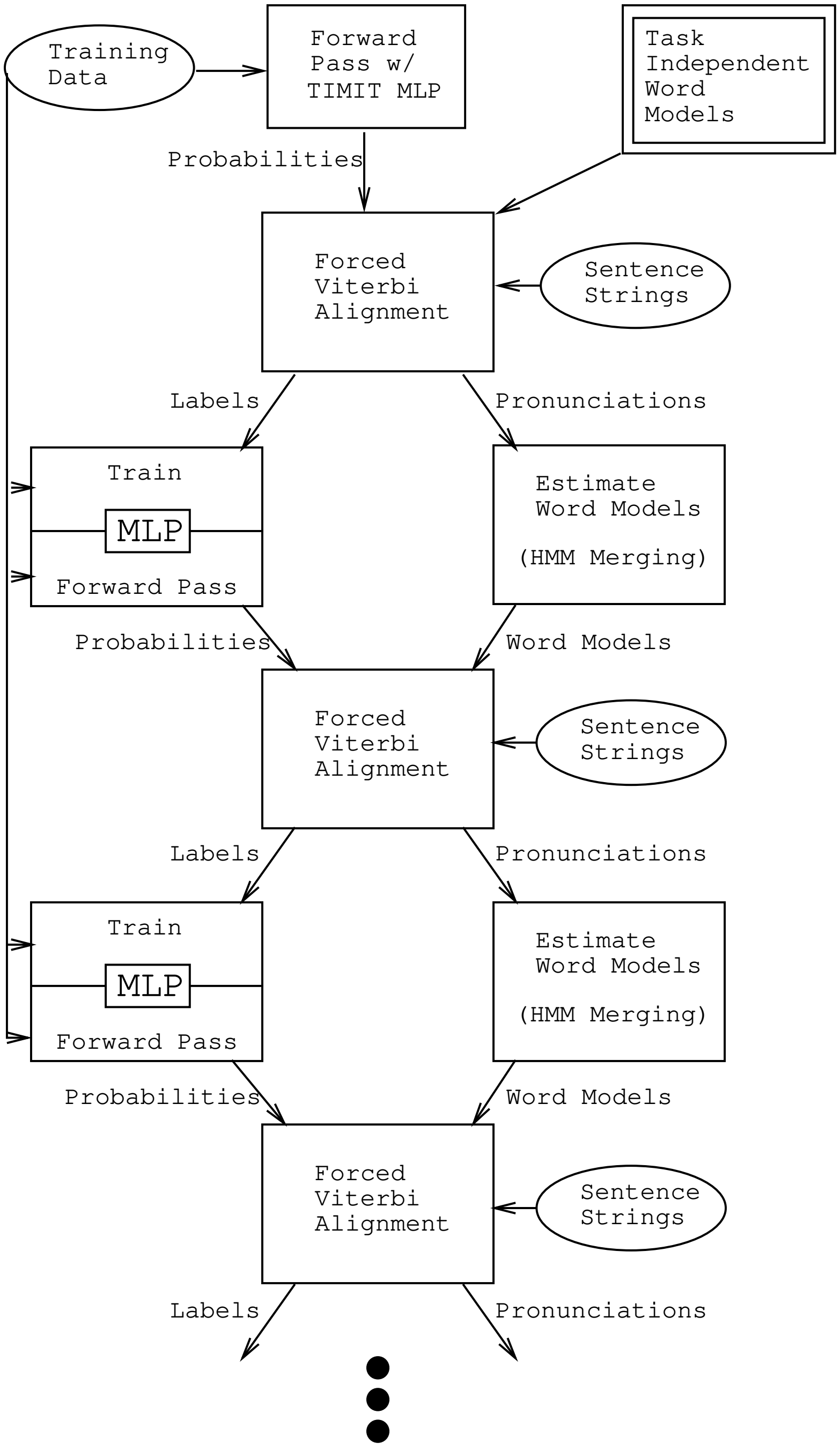

Figure 18: Hybrid MLP/HMM training/merging procedure used in the BeRP speech understanding system (Wooters 1993).

To assess its effectiveness, the recognition performance of the multiple-pronunciation system was compared against that of an otherwise identical system in which only one phone sequence per word was used (generated by a commercial text-to-speech system). In this comparison, multiple-pronunciation modeling as derived by HMM merging was found to reduce the word-level error rate from 40.6% to 32.1%. At the same time, the error rate at the level of semantic interpretations dropped from 43.4% to 34.1%.

Further experiments are needed to identify more precisely what aspects of the multiple-pronunciation approach account for the improvement, *i.e.*, whether other word model building techniques would lead to significantly different results in this context. However, the preliminary results do show that HMM merging is both practical and effective when embedded in a realistic speech system.

The details of the construction of these word models, along with discussion of ancillary issues and a graphical HMM representation of the pronunciation for the 50 most common words in the BeRP corpus can be found in Wooters (1993).

# 7 Conclusions and Further Research

Our evaluations indicate that the HMM merging approach is a promising new way to induce probabilistic finite-state models from data. It compares favorably with the standard Baum-Welch method, especially when there are few prior constraints on the HMM topology. Our implementation of the algorithm and applications in the speech domain have shown it to be feasible in practice.

Experimentation with the range of plausible priors, as well as new, application-specific ones is time-consuming, and we have barely scratched the surface in this area. However, the experience so far with the priors discussed in Section 3.4 is that the particular choice of prior type and parameters does not greatly affect the course of the best-first search, except possibly the decision when to stop. In other words, the merging heuristic, together with the effect of the likelihood are the determining factors in the choice of merges. This could, and in fact should, change with the use of more informative priors.

Likewise, we haven't pursued merging of HMMs with non-discrete outputs. For example, HMMs with mixtures of Gaussians as emission densities are being used extensively (Gauvain & Lee 1991) for speech modeling. Our merging algorithm becomes applicable to such models provided that one has a prior for such densities, which should be straightforward (Cheeseman *et al.* 1988). Efficient implementation of the merging operator may be a bigger problem—one wants to avoid having to explicitly compute a merged density for each merge under consideration.

One of the major shortcomings of the current merging strategy is its inability to 'back off' from a merging step that turns out be an overgeneralization in the light of new data. A solution to this problem might be the addition of a complementary state *splitting* operator, possibly along the lines of Ron *et al.* (1994). The major difficulty with evaluating splits (as opposed to merges) is that it requires rather more elaborate statistics than simple Viterbi counts, since splitting decisions are based on co-occurrences of states in a path.

Our current work focuses on induction of richer grammatical models, especially stochastic context-free grammars, within the same Bayesian model merging framework. The structure finding problem in this domain is even more severe, as standard EM-based estimation methods have great difficulty when presented with unstructured, fully parameterized grammars (Lari & Young 1990; Pereira & Schabes 1992). We have achieved good results inducing SCFGs using a merging heuristic that is a direct generalization of the one used for HMMs; this approach will be described in a forthcoming publication.

## Acknowledgements

Gary Tajchman conducted first experiments on the TIMIT data using our algorithm. Chuck Wooters carried out all recognition experiments involving TIMIT and BeRP acoustic data. Besides providing us with valuable results, both contributed many suggestions and helped make the implementation more flexible and robust.

We would like to thank Peter Cheeseman, Wray Buntine, David Stoutamire, Jerry Feldman, and Fernando Pereira for helpful discussions of the issues in this paper, as well as numerous members of the ICSI AI and Speech groups for valuable comments on drafts.

## References

ANGLUIN, D., & C. H. SMITH. 1983. Inductive inference: Theory and methods. *ACM Computing Surveys* 15.237–269.

BALDI, PIERRE, YVES CHAUVIN, TIM HUNKAPILLER, & MARCELLA A. MCCLURE. 1993. Hidden Markov models in molecular biology: New algorithms and applications. In *Advances in Neural Information Processing Systems 5*, ed. by Stephen José Hanson, Jack D. Cowan, & C. Lee Giles, 747–754. San Mateo, CA: Morgan Kaufmann.

BAUM, LEONARD E., TED PETRIE, GEORGE SOULES, & NORMAN WEISS. 1970. A maximization technique occuring in the statistical analysis of probabilistic functions in Markov chains. *The Annals of Mathematical Statistics* 41.164–171.

BOURLARD, HERVÉ, & NELSON MORGAN. 1993. *Connectionist Speech Recognition. A Hybrid Approach.* Boston, Mass.: Kluwer Academic Publishers.

BROWN, PETER F., VINCENT J. DELLA PIETRA, PETER V. DESOUZA, JENIFER C. LAI, & ROBERT L. MERCER. 1992. Class-based $n$-gram models of natural language. *Computational Linguistics* 18.467–479.

BUNTINE, W. L. 1991. Theory refinement of Bayesian networks. In *Seventh Conference on Uncertainty in Artificial Intelligence*, Anaheim, CA.

BUNTINE, WRAY. 1992. Learning classification trees. In *Artificial Intelligence Frontiers in Statistics: AI and Statistics III*, ed. by D. J. Hand. Chapman & Hall.

CHEESEMAN, PETER, JAMES KELLY, MATTHEW SELF, JOHN STUTZ, WILL TAYLOR, & DON FREEMAN. 1988. AutoClass: A Bayesian classification system. In *Proceedings of the 5th International Conference on Machine Learning*, 54–64, University of Michigan, Ann Arbor, Mich.

CHEN, FRANCINE R. 1990. Identification of contextual factors for pronunciation networks. In *Proceedings IEEE Conference on Acoustics, Speech and Signal Processing*, volume 2, 753–756, Albuquerque, NM.

CLEEREMANS, AXEL, 1991. *Mechanisms of Implicit Learning. A Parallel Distributed Processing Model of Sequence Acquisition.* Pittsburgh, Pa.: Department of Psychology, Carnegie Mellon University dissertation.

COOPER, GREGORY F., & EDWARD HERSKOVITS. 1992. A Bayesian method for the induction of probabilistic networks from data. *Machine Learning* 9.309–347.

COVER, THOMAS M., & JOY A. THOMAS. 1991. *Elements of Information Theory.* New York: John Wiley and Sons, Inc.

CUTTING, DOUG, JULIAN KUPIEC, JAN PEDERSEN, & PENELOPE SIBUN. 1992. A practical part-of-speech tagger. In *Proceedings of the Third Conference on Applied Natural Language Processing*, Trento, Italy. ACL. Also available as Xerox PARC Technical Report SSL-92-01.

DEMPSTER, A. P., N. M. LAIRD, & D. B. RUBIN. 1977. Maximum likelihood from incomplete data via the *EM* algorithm. *Journal of the Royal Statistical Society, Series B* 34.1–38.

GAROFOLO, J. S., 1988. *Getting Started with the DARPA TIMIT CD-ROM: an Acoustic Phonetic Continuous Speech Database*. National Institute of Standards and Technology (NIST), Gaithersburgh, Maryland.

GAUVAIN, JEAN-LUC, & CHIN-HIN LEE. 1991. Bayesian learning of Gaussian mixture densities for hidden Markov models. In *Proceedings DARPA Speech and Natural Language Processing Workshop*, 271–277. Pacific Grove, CA: Defence Advanced Research Projects Agency, Information Science and Technology Office.

GULL, S. F. 1988. Bayesian inductive inference and maximum entropy. In *Maximum Entropy and Bayesian Methods in Science and Engineering, Volume 1: Foundations*, ed. by G. J. Erickson & C. R. Smith, 53–74. Dordrecht: Kluwer.

HAUSSLER, DAVID, ANDERS KROGH, I. SAIRA MIAN, & KIMMEN SJÖLANDER. 1992. Protein modeling using hidden Markov models: Analysis of globins. Technical Report UCSC-CRL-92-23, Computer and Information Sciences, University of California, Santa Cruz, Ca. Revised Sept. 1992.

HOPCROFT, JOHN E., & JEFFREY D. ULLMAN. 1979. *Introduction to Automata Theory, Languages, and Computation*. Reading, Mass.: Addison-Wesley.

HORNING, JAMES JAY. 1969. A study of grammatical inference. Technical Report CS 139, Computer Science Department, Stanford University, Stanford, Ca.

JELINEK, FREDERICK, & ROBERT L. MERCER. 1980. Interpolated estimation of Markov source parameters from sparse data. In *Proceedings Workshop on Pattern Recognition in Practice*, 381–397, Amsterdam.

JURAFSKY, DANIEL, CHUCK WOOTERS, GARY TAJCHMAN, JONATHAN SEGAL, ANDREAS STOLCKE, & NELSON MORGAN. 1994. The Berkeley Restaurant Project. Technical report, International Computer Science Institute, Berkeley, CA. To appear.

KATZ, SLAVA M. 1987. Estimation of probabilities from sparse data for the language model component of a speech recognizer. *IEEE Transactions on Acoustics, Speech, and Signal Processing* 35.400–401.

LARI, K., & S. J. YOUNG. 1990. The estimation of stochastic context-free grammars using the Inside-Outside algorithm. *Computer Speech and Language* 4.35–56.

OMOHUNDRO, STEPHEN M. 1992. Best-first model merging for dynamic learning and recognition. In *Advances in Neural Information Processing Systems 4*, ed. by John E. Moody, Steve J. Hanson, & Richard P. Lippman, 958–965. San Mateo, CA: Morgan Kaufmann.

Pereira, Fernando, & Yves Schabes. 1992. Inside-outside reestimation from partially bracketed corpora. In *Proceedings of the 30th Annual Meeting of the Association for Computational Linguistics*, 128–135, University of Delaware, Newark, Delaware.

Porat, Sara, & Jerome A. Feldman. 1991. Learning automata from ordered examples. *Machine Learning* 7.109–138.

Quinlan, J. Ross, & Ronald L. Rivest. 1989. Inferring decision trees using the minimum description length principle. *Information and Computation* 80.227–248.

Rabiner, L. R., & B. H. Juang. 1986. An introduction to hidden Markov models. *IEEE ASSP Magazine* 3.4–16.

Reber, A. S. 1969. Implicit learning of artifical grammars. *Journal of Verbal Learning and Verbal Behavior* 6.855–863.

Redner, Richard A., & Homer F. Walker. 1984. Mixture densities, maximum likelihood and the EM algorithm. *SIAM Review* 26.195–239.

Riley, Michael D. 1991. A statistical model for generating pronunciation networks. In *Proceedings IEEE Conference on Acoustics, Speech and Signal Processing*, volume 2, 737–740, Toronto.

Rissanen, Jorma. 1983. A universal prior for integers and estimation by minimum description length. *The Annals of Statistics* 11.416–431.

Ron, Dana, Yoram Singer, & Naftali Tishby. 1994. The power of amnesia. In *Advances in Neural Information Processing Systems 6*, ed. by Jack Cowan, Gerald Tesauro, & Joshua Alspector. San Mateo, CA: Morgan Kaufmann.

Stolcke, Andreas, & Stephen Omohundro. 1993. Hidden Markov model induction by Bayesian model merging. In *Advances in Neural Information Processing Systems 5*, ed. by Stephen José Hanson, Jack D. Cowan, & C. Lee Giles, 11–18. San Mateo, CA: Morgan Kaufmann.

Thomason, Michael G., & Erik Granum. 1986. Dynamic programming inference of Markov networks from finite set of sample strings. *IEEE Transactions on Pattern Analysis and Machine Intelligence* 8.491–501.

Tomita, Masaru. 1982. Dynamic construction of finite automata from examples using hill-climbing. In *Proceedings of the 4th Annual Conference of the Cognitive Science Society*, 105–108, Ann Arbor, Mich.

Viterbi, A. 1967. Error bounds for convolutional codes and an asymptotically optimum decodning algorithm. *IEEE Transactions on Information Theory* 260–269.

Wallace, C. S., & P. R. Freeman. 1987. Estimation and inference by compact coding. *Journal of the Royal Statistical Society, Series B* 49.240–265.

Wooters, Charles C., 1993. *Lexical Modeling in a Speaker Independent Speech Understanding System*. Berkeley, CA: University of California dissertation.